\documentclass[lettersize,journal]{IEEEtran}
\usepackage{amsmath,amsfonts}
\usepackage{algorithmic}
\usepackage{algorithm}
\usepackage{array}
\usepackage[caption=false,font=normalsize,labelfont=sf,textfont=sf]{subfig}
\usepackage{textcomp}
\usepackage{stfloats}
\usepackage{url}
\usepackage{verbatim}
\usepackage{graphicx}
\usepackage{cite}
\usepackage{cleveref}
\usepackage{booktabs}
\usepackage{multirow}
\Crefname{figure}{Fig.}{Figs.}     
\crefname{algorithm}{Algorithm}{Algorithms}

\begin{document}

\title{Adversarial Flow Matching for Imperceptible Attacks on End-to-End Autonomous Driving}

\author{Xinyu Zeng, 
Xiangkun He, \IEEEmembership{Senior Member, IEEE},
	Lei Tao, 
	Chen Lv, \IEEEmembership{Senior Member, IEEE},
    Hong Cheng, \IEEEmembership{Senior Member, IEEE}
\thanks{Xinyu Zeng, Xiangkun He, and Lei Tao are with the Shenzhen Institute for Advanced Study, University of Electronic Science and Technology of China, Shenzhen 518110, China (e-mail: 202422280324@std.uestc.edu.cn; xiangkun.he@uestc.edu.cn; 202522280735@std.uestc.edu.cn). \emph{(Corresponding author: Xiangkun He.)}}
\thanks{Chen Lv is with the School of Mechanical and Aerospace Engineering, Nanyang Technological University, Singapore 639798 (e-mail: lyuchen@ntu.edu.sg).}
\thanks{Hong Cheng is with the School of Mechanical and Electrical Engineering, University of Electronic Science and Technology of
China, Chengdu 611731, China (e-mail: hcheng@uestc.edu.cn).}

}



\maketitle

\begin{abstract}
Autonomous driving (AD) is evolving towards end-to-end (E2E) frameworks through two primary paradigms: monolithic models exemplified by Vision-Language-Action (VLA), and specialized modular architectures. Despite their divergent designs, both paradigms increasingly rely on Transformer backbones for complex reasoning, potentially causing a shared vulnerability: visually imperceptible perturbations can manipulate E2E AD models into hazardous maneuvers by targeting the Transformer module. 
Most existing adversarial attack approaches against AD systems operate under white-box or black-box settings; yet, they typically necessitate full model transparency, or suffer from either prohibitive query latency or limited attack transferability. 
In this paper, we propose Adversarial Flow Matching (AFM), a novel gray-box attack framework that exploits Transformer structural vulnerabilities in E2E AD models. 
AFM enables efficient one-step generation of adversarial examples via a neural average velocity field. 
Additionally, the proposed technique yields effective and visually imperceptible attacks by synergistically perturbing the generative latent space and the neural average velocity field. Extensive experiments demonstrate that AFM achieves a superior trade-off between attack effectiveness and imperceptibility: it substantially degrades the performance of both VLA and modular AD agents across various scenarios compared to baselines, while maintaining state-of-the-art visual imperceptibility. 
Furthermore, adversarial examples generated by AFM exhibit robust cross-model transferability, indicating that AFM closely approximates a black-box attack setting while requiring only the prior knowledge that the target AD model incorporates a Transformer-based module.
\end{abstract}

\begin{IEEEkeywords}
End-to-end autonomous driving, vision-language-action model, adversarial attack, flow matching.
\end{IEEEkeywords}

\begin{figure}[h]
    \vspace{-8mm}  
    \centering
    \hspace*{-0.25cm}
    \includegraphics[width=0.5\textwidth]{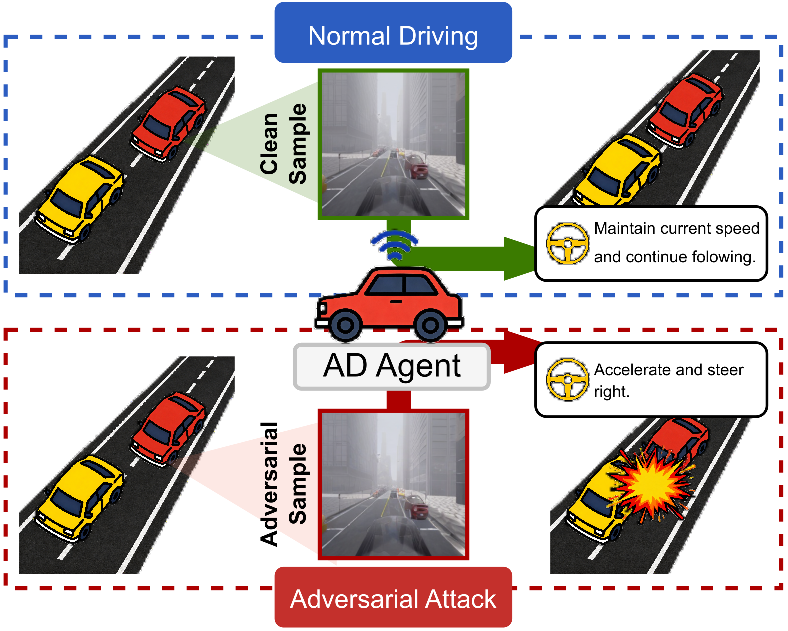}
    \caption{\emph{Illustration of AD behaviors under normal driving situations and adversarial attacks. } 
   Top: An autonomous vehicle functions as an AD agent that interacts with its environment through a continuous "perception-action" loop, where clean sensory samples result in safe and stable trajectories.
   Bottom: An adversarial attack disrupts this closed-loop cycle by injecting imperceptible perturbations; the resulting adversarial samples mislead the agent into hazardous decision-making and unsafe maneuvers, such as abrupt acceleration or erroneous steering.}
    \label{fig:abstract}
    \vspace{-5mm}  
\end{figure}

\section{Introduction}
\IEEEPARstart{D}{riven} by rapid breakthroughs in foundation models, autonomous driving (AD) systems are transitioning from traditional modular designs to end-to-end (E2E) architectures, evolving into two dominant paradigms: monolithic architectures exemplified by Vision-Language-Action (VLA) models \cite{boroujeni2026vla4codrive, jiang2025survey}, and specialized modular architectures \cite{chen2024end}. 

Despite differing in their underlying frameworks, both paradigms incorporate Transformer backbones to bolster complex reasoning, which simultaneously inherit the canonical fragilities of deep neural networks \cite{szegedy2013intriguing}.
Typically, an AD vehicle operates by continuously perceiving its surroundings and executing maneuvers in response to dynamic environmental changes. This closed-loop interaction effectively characterizes the vehicle as an AD agent that pursues goal-directed tasks in complex traffic scenarios.
As illustrated in \Cref{fig:abstract}, even visually imperceptible perturbations injected into clean samples to craft adversarial samples, can induce hazardous driving behaviors. The growing interconnectivity of AD agents further exacerbates this risk, in practical driving environments, threats ranging from physical adversarial patches on roadside infrastructure to digital noise injected via Vehicle-to-Everything (V2X) networks can trigger unforeseen and catastrophic safety incidents.
Consequently, it is of paramount importance to proactively identify the inherent vulnerabilities of modern AD agents, particularly emerging monolithic models exemplified by VLA. 
Current research on adversarial attacks targeting AD agents is predominantly dichotomized into white-box and black-box paradigms. White-box attacks \cite{zhang2024uniada,zhang2024visual,wang2024attack,xu2025challenger,wang2023driving,zheng2024physical,hu2025dynamicpae} typically achieve superior efficacy by leveraging full access to the agent's internal gradients and structures; however, they are severely constrained in real-world deployment due to the assumption of complete agent transparency. In contrast, black-box attacks on AD agents generally bifurcate into query-based and transfer-based paradigms. Query-based attacks \cite{9425267,255240} are incompatible with the real-time requirements of  AD agents due to the prohibitive latency incurred by extensive interactions, while transfer-based attacks \cite{279980,wang2025black,boloor2020attacking} suffer from performance degradation stemming from the misalignment between the surrogate and the target driving policy. 
Crucially, gray-box attacks remain underexplored in the AD agents domain. Unlike white-box attacks or black-box attacks, gray-box attacks leverage limited but vital information. For instance, gray-box attacks may exploit structural knowledge of the AD agent, such as its structure backbone or sensor fusion protocols, yet they remain blind to the specific model weights or the exact cost functions that govern trajectory planning.

While conventional attacks predominantly rely on pixel-wise additive perturbations, the field is witnessing a shift toward generative adversarial paradigms, aiming to enhance sample diversity and visual imperceptibility. 
By capturing multi-modal distributions and latent semantic manifolds, these approaches synthesize diverse and visually natural perturbations. 
Capitalizing on these inherent properties,
recent frameworks utilizing GANs \cite{kong2020physgan,fan2024adversarial} and Diffusion Models \cite{xu2025challenger,zhao2025generating} have demonstrated the capability to generate adversarial examples that are both effective and visually naturalistic.
In this work, we propose Adversarial Flow Matching (AFM), an imperceptible gray-box attack 
designed to uncover inherent vulnerabilities within AD agents, particularly those manifests in VLA-driven paradigms. 
The central idea lies in a dual-perturbation paradigm that injects learnable perturbations into both the latent space and the neural average velocity field \cite{geng2025mean}. By exploiting Transformer-specific gradients, we synergistically optimize these two components to steer the generative trajectory toward adversarial manifolds. This mechanism augments latent space attacks with velocity field perturbations, thereby potentially inducing perturbation to the latent semantic features of Transformer-based AD agents while strictly preserving visual fidelity. Furthermore, by capitalizing on the intrinsic straight-path property of Flow Matching trajectories, AFM facilitates direct one-step (1-NFE) adversarial generation, effectively circumventing the computational overhead and accumulated discretization errors that typically plague iterative generative attack frameworks. We conduct extensive empirical evaluations including both open-loop dataset evaluations and highly challenging closed-loop interactive simulations (e.g., CARLA) demonstrating that AFM substantially degrades the performance of both VLA and modular AD agents in various scenarios, outperforming existing baselines. Furthermore, adversarial examples generated by AFM exhibit robust cross-model transferability. 
Unlike studies dedicated to defense, our research focuses on investigating a novel adversarial attack framework. 
By requiring only a structural prior of the Transformer backbone,we aim to investigate the feasibility of efficient and visually imperceptible adversarial attacks against AD agents. Such capabilities provide critical diagnostic insights into model vulnerabilities, facilitating the development of next-generation, inherently robust systems—specifically VLA-driven AD agents.
Our main contributions are as follows:
\begin{itemize}
    \item We propose Adversarial Flow Matching (AFM), a novel gray-box attack framework designed as a security benchmark to expose the critical structural vulnerabilities inherent to the Transformer backbones of AD agents.
    \item To our knowledge, AFM represents the first attempt to leverage Flow Matching for adversarial example generation against AD agents,achieving one-step (1-NFE) generation to effectively bypass the iterative sampling latency of traditional generative attacks via a neural average velocity field.
    \item We validate AFM across both monolithic VLA architectures and specialized modular architectures, where it exhibits a favorable balance of visual imperceptibility, attack efficacy, and generation efficiency.
    \item Our cross-model transferability experiments consistently demonstrate that AFM is capable of executing effective attacks with the sole knowledge that the target AD agent incorporates a Transformer-based module. This structural prior allows AFM to achieve high attack efficacy while remaining visually imperceptible in realistic driving scenarios, effectively bypassing the need for internal access to the AD agent.
\end{itemize}

\section{Related Work}
\subsection{\textbf{Adversarial Attacks on AD agents}}
With the widespread deployment and remarkable performance of AD agents \cite{zhao2024autonomous,levinson2011towards}, their robustness has emerged as a critical concern \cite{gao2021autonomous,ibrahum2024deep,jindi2022evaluating}. However, both monolithic architectures and modular multi-model frameworks inherently inherit the vulnerabilities of Deep Neural Networks \cite{szegedy2013intriguing}. Given that even visually imperceptible perturbations can induce dangerous maneuvers \cite{sobh2021adversarial,deng2021deep}, adversarial attacks on AD agents have garnered significant research attention.

For white-box scenarios, several recent studies have explored adversarial strategies adapted for end-to-end AD agents.
UniAda \cite{zhang2024uniada} leverages adaptive weighting scheme and multi-objective optimization to simultaneously disrupt both longitudinal and lateral controls, demonstrating that AD agents are highly vulnerable to synchronized multi-dimensional attacks. Module-Wise Attack \cite{wang2024attack} injects adversarial noise at sub-module interfaces,leveraging error accumulation to induce severe planning deviations. However, these approaches are predicated on the assumption of full access to the target model’s architecture and parameters; such a prerequisite of complete transparency renders them impractical for real-world deployments. For the emerging VLA-based AD agents, white-box attacks now focus on semantic consistency across temporal sequences. Research such as ADvLM \cite{zhang2024visual} exploits the cross-modal attention mechanisms to optimize perturbations that remain effective under varied textual instructions and continuous visual frames,but suffer from high computational overhead.

Black-box attacks present a significant practical threat by precluding the requirement for internal model accessibility. M-SimBA \cite{9425267} improves convergence efficiency by targeting the most confused class, while \cite{255240} leverages inherent geometric occlusions in LiDAR point clouds to ensure physical stealth. However, both methodologies are characterized by high query complexity; this substantial overhead restricts their efficacy in real-time driving scenarios while increasing their susceptibility to defenses.
Recent work like CAD \cite{wang2025black} shifts the focus from pixel-level noise to logical reasoning disruption. By leveraging surrogate models to inject deceptive semantics, it can paralyze the perception-to-reasoning chain in VLA-based AD agents. Nevertheless, its practical efficacy is often hampered by the transferability gap between surrogate and target models, compounded by significant computational overhead.

While gray-box attacks represent a plausible and potent threat in real-world scenarios by assuming only partial system knowledge, they remain notably underexplored within the AD agents domain, particularly in the context of emerging VLA-based agents. To reconcile the stringent requirements for full model transparency in white-box attacks with the high query complexity and limited transferability of black-box attacks, we propose a novel gray-box framework specifically tailored for the Transformer backbone, which enables complex reasoning for both monolithic VLA and specialized modular architectures. By capitalizing on the intrinsic structural vulnerabilities of this shared foundation, our method strikes a favorable balance among attack efficacy, computational efficiency, and practical deployability.
\subsection{\textbf{Imperceptible Attack}}
Balancing adversarial efficacy with visual imperceptibility remains a pivotal challenge. To transcend the artifacts inherent in conventional Lp-norm constraints, researchers have integrated Human Visual System (HVS) \cite{luo2022frequency} characteristics into the optimization process. For instance, PerC-AL \cite{Zhao_2020_CVPR} and NCF \cite{yuan2022natural} leverage physiological color perception to disguise perturbations as natural illumination or chrominance shifts, while SSAH \cite{luo2022frequency} exploit spectral insensitivity to embed disruptions in specific frequency bands. While these methods exhibit superior visual imperceptibility and robustness against image-processing defenses, their efficacy is often contingent upon precise semantic priors or specific spectral signatures, which incur substantial optimization overhead. 
To overcome these limitations, generative paradigms have emerged as a promising alternative, utilizing probabilistic models to align adversarial examples with the natural data distribution and facilitate diverse sample generation.
AdvINN \cite{Chen_Wang_Huang_Zhao_Liu_Guan_2023} and AdvFlow \cite{mohaghegh2020advflow} utilize Invertible Neural Networks and Normalizing Flows to manipulate latent representations, ensuring that perturbations are embedded within the data manifold. Building upon this, DiffAttack \cite{kang2023diffattack} introduces diffusion models to the adversarial field, exploiting their powerful generative and discriminative priors to craft visual imperceptive perturbations rich in semantic clues. Despite boosting black-box transferability, these generative approaches are fundamentally limited by high sampling latency and computational intensity. Notably, the pursuit of adversarial imperceptibility has also extended to the specific context of AD. PhysGAN \cite{kong2020physgan} and Challenger \cite{xu2025challenger} focus on environmental and temporal consistency, utilizing generative frameworks to maintain adversarial potency across varying viewpoints or coherent video streams. Trajectory Attack \cite{fan2024adversarial} and Diffusion Point Cloud \cite{zhao2025generating} leveraging GANs and diffusion priors to disrupt trajectory forecasting and LiDAR-based geometric distortions. While these methods effectively expose vulnerabilities across the prediction-to-planning pipeline, they often incur substantial computational overhead due to iterative sampling and spatio-temporal optimization. This limits their generation efficiency, presenting challenges for deployment in dynamic, real-world driving scenarios.

In this work, we introduce Flow Matching to the domain of adversarial attacks in AD agents. By leveraging the intrinsic straightness of learned Optimal Transport (OT) paths inherent to this paradigm, our framework enables efficient one-step adversarial generation by injecting disturbances into a neural velocity field.
This approach significantly mitigates the discretization errors typical of curved trajectories, allowing our method to simultaneously bypass iterative optimization latency to facilitate efficient adversarial generation, while maintaining high visual imperceptibility.

\begin{figure*}[h]
    \centering
    \vspace{-15pt}
    \includegraphics[width=0.95\textwidth]{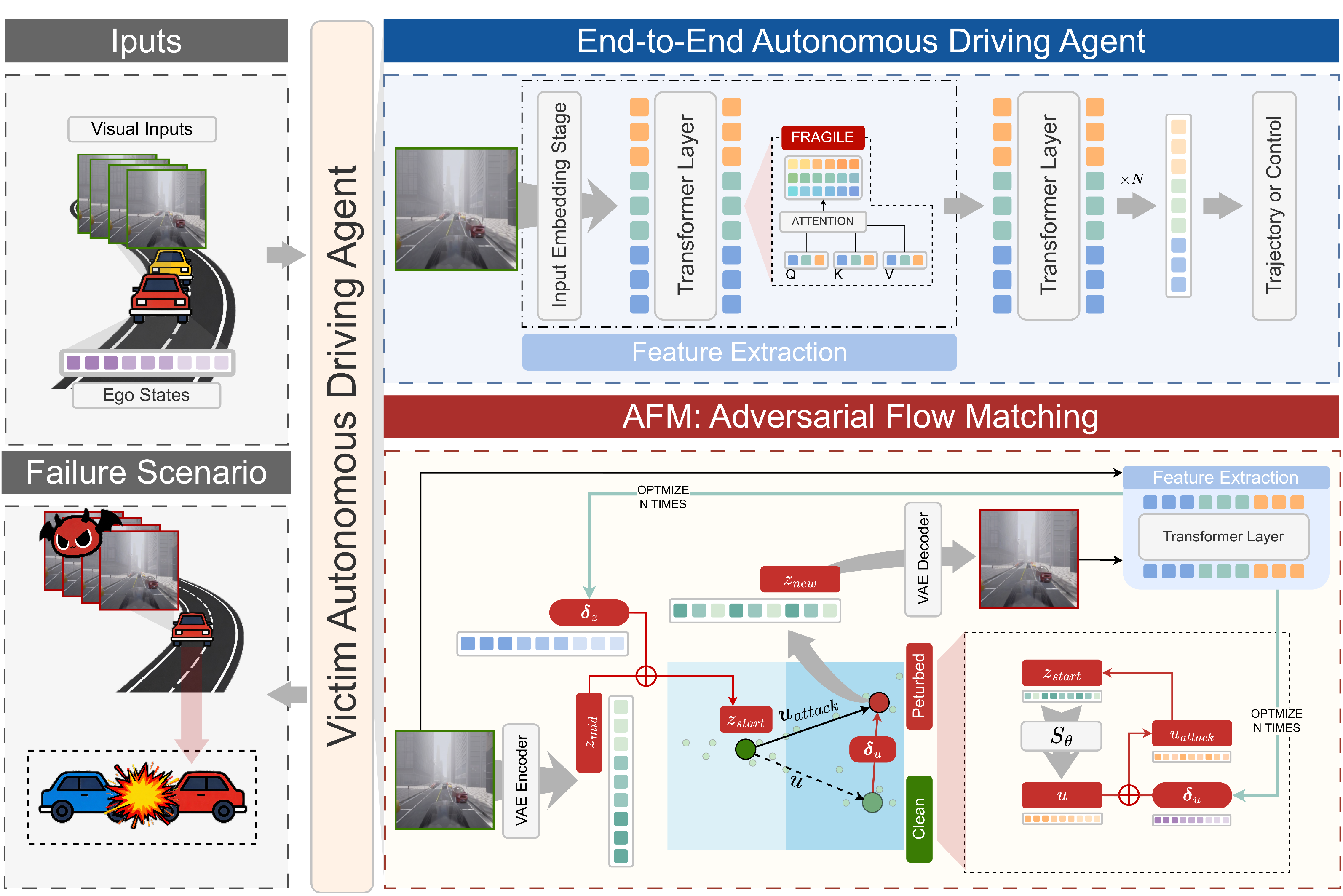}
    \vspace{-8pt}
    \caption{\emph{Overview of the proposed AFM framework.} The upper part illustrates a victim end-to-end AD agent, where visual inputs and ego states are encoded and processed by a Transformer-based perception and planning pipeline to generate trajectories or control commands. The lower part depicts the AFM attack mechanism. Given a clean sample, AFM performs a one-step inversion into the latent space using a frozen VAE encoder, and introduces dual perturbations on both the latent initialization and the neural average velocity field. By jointly manipulating the latent trajectory and the flow field, AFM generates visually imperceptible adversarial perturbations that potentially induce hazardous driving behaviors, while keeping all victim model parameters fixed.}
    \label{fig:afm_framework}
    \vspace{-5pt}
\end{figure*}

\section{Methodology}
In this section, we introduce the proposed attack framework and elaborate on the technical specifics.
\subsection{\textbf{Problem Definition}}
Let $\mathcal{M} : \mathcal{X} \rightarrow \mathcal{Y}$ denote an end-to-end autonomous driving agent that maps sensory inputs
$\mathbf{x} \in \mathbb{R}^{H \times W \times 3}$ to planning trajectories
$\mathbf{y} \in \mathbb{R}^{T \times 2}$.
Consistent with modern foundation model architectures for AD agents, $\mathcal{M}$ incorporates Transformer backbones.

We denote the Transformer backbone as $\Phi(\cdot)$ and the downstream reasoning module as $\varphi(\cdot)$.
The inference process is formulated as:
\begin{equation}
\mathbf{y} = \varphi(\Phi(\mathbf{x}), \mathbf{c}),
\end{equation}
where $\mathcal{F} = \Phi(\mathbf{x})$ represents the extracted feature embeddings and
$\mathbf{c}$ denotes auxiliary information, such as navigation commands.

The objective of an adversary is to craft a perturbation that deviates the model's decision.
Formally, the adversarial output is given by:
\begin{equation}
\mathbf{y}' = \varphi(\Phi(\mathbf{x}_{adv}), \mathbf{c}),
\quad \text{s.t.} \quad \lVert \mathbf{y}' - \mathbf{y} \rVert > \varepsilon,
\end{equation}
where $\mathbf{x}_{adv}$ is the synthesized adversarial example and $\varepsilon$ is a predefined deviation threshold.

\subsubsection{\textbf{Threat Model}}
We consider a gray-box adversarial attack setting. In this scenario, the adversary lacks access to the full internal parameters of the target model (specifically $\varphi$). However, we assume access to the gradients of the Transformer backbone $\Phi(\cdot)$ is available. This allows the adversary to backpropagate gradients through $\Phi(\cdot)$ to craft adversarial examples in the input space.

\subsubsection{\textbf{Optimization Objective}}
Given a clean sample $\mathbf{x}_{clean}$, our goal is to synthesize an adversarial example
$\mathbf{x}_{adv}$ via a generative process.
The optimization objective is twofold: 

\emph{(i) Feature Distortion.} Maximize the deviation of the latent semantic features $\mathcal{F} = \Phi(\mathbf{x})$ within task-critical regions, thereby inducing the downstream reasoning module $\varphi(\cdot)$ to generate erroneous predictions; 

\emph{(ii) Visual Fidelity Preservation.} Ensure that $\mathbf{x}_{adv}$ strictly maintains visual imperceptibility and that the underlying latent statistics of the generative model remain stable.

\subsubsection{\textbf{Attack Formulation}}
We formulate the attack within the latent space and the neural average velocity field.
Let $\mathcal{S}_{1 \to 0}(\cdot)$ denote the Flow Matching Sampler (Generation), and $\mathcal{S}_{0 \to 1}(\cdot)$ denote the inversion operation mapping the clean latent representation back into the noise space by reversing the deterministic sampling process.
We optimize both a latent perturbation $\boldsymbol{\delta}_z$ and a neural average velocity field perturbation $\boldsymbol{\delta}_u$. Formally, the attack is defined as the following constrained optimization problem:
\begin{equation}
\begin{aligned}
\max_{\boldsymbol{\delta}_z,\,\boldsymbol{\delta}_u} \quad
& \mathcal{L}_{attack}\!\left(
\Phi(\mathbf{x}_{adv}), \,
\Phi(\mathbf{x}_{clean}); \,
\mathbf{M}, \mathbf{A}
\right) \\
\text{s.t.} \quad
& \mathbf{z}_{clean} = \mathcal{E}(\mathbf{x}_{clean}), \\
& \mathbf{z}_{mid} =\mathcal{S}_{0 \to 1}(\mathbf{z}_{clean}), \\
& \mathbf{x}_{adv} = \mathcal{D}(\mathcal{S}_{1 \to 0}(\mathbf{z}_{mid} + \boldsymbol{\delta}_z; \boldsymbol{\delta}_u)), \\
& \lVert \boldsymbol{\delta}_z \rVert_{\infty} \le \epsilon_z, \quad
  \lVert \boldsymbol{\delta}_u \rVert_{\infty} \le \epsilon_u, \\
& \left| \sigma(\mathbf{z}_{adv}) - \sigma(\mathbf{z}_{clean}) \right| \le \xi ,
\end{aligned}
\end{equation}
where $\sigma(\cdot)$ denotes the standard deviation of latent variables, serving to enforce statistical consistency and prevent generative collapse. $\mathcal{E}(\cdot)$ and $\mathcal{D}(\cdot)$ denote the encoder and decoder of the pre-trained VAE, respectively. $\mathbf{M}$ and $\mathbf{A}$ represent the spatial mask and attention map used for weighting the loss.

\subsection{\textbf{Adversarial Flow Matching}}
As illustrated in \Cref{fig:afm_framework}, our proposed attack framework comprises two key components: \emph{(i) a Flow Matching--guided generative mechanism} enabling efficient single-step adversarial generation, and \emph{(ii) an attention-guided multi-objective optimization} tailored for the decision dynamics of AD agents. 

The algorithmic procedure of our proposed AFM is formalized in \cref{alg:afm}, it outlines the step-by-step procedure of AFM, formally detailing the initialization, dual-perturbation injection into the latent space and velocity field, and the gradient-based optimization process.

\begin{algorithm}[htbp]
  \caption{AFM: Adversarial Flow Matching}
  \label{alg:afm}
  \begin{algorithmic}[1]
    \REQUIRE Clean image $\mathbf{x}_{clean}$; 
    Target vision backbone $\Phi$ (frozen);
    Flow matching network $\mathcal{S}_\theta$ (frozen);
    VAE encoder $\mathcal{E}$ / decoder $\mathcal{D}$.
    \REQUIRE Hyperparameters: steps $N$; rates $\eta_z, \eta_u$;
    bounds $\epsilon_z, \epsilon_u$; weights $\lambda_f, \lambda_a, \lambda_{c}$; temp $\tau$.
    \ENSURE Adversarial example $\mathbf{x}_{adv}$.
    \STATE \textbf{Initialization:}
    \STATE $\mathbf{z}_{clean} \gets \mathcal{E}(\mathbf{x}_{clean})$
    \STATE $\mathbf{z}_{mid} \gets \mathcal{S}_{0 \to 1}(\mathbf{z}_{clean})$ \COMMENT{Invert flow to noise space}
    \STATE $\mathcal{F}_{clean}, \mathbf{A}_{clean} \gets \Phi(\mathbf{x}_{clean})$
    \STATE $\boldsymbol{\delta}_z \gets \mathbf{0}, \quad \boldsymbol{\delta}_u \gets \mathbf{0}$
    \STATE $\mathbf{M} \gets \text{SpatialMask}(\text{Road})$ \COMMENT{Construct spatial mask}
    \STATE $\mathbf{W}_{attn} \gets \text{Softmax}(\mathbf{A}_{clean}/\tau)$ \COMMENT{Pre-compute attn weights}
    \STATE \textbf{Optimization Loop:}
    \FOR{$n=1$ {\bfseries to} $N$}
      \STATE \textit{(1) Perturb latent and velocity}
      \STATE $\mathbf{z}_{start} \gets \mathbf{z}_{mid} + \boldsymbol{\delta}_z$
      \STATE $\mathbf{u} \gets \mathcal{S}_\theta(\mathbf{z}_{start}, t_{start}, t_{end})$
      \STATE $\mathbf{u}_{attack} \gets \mathbf{u} + \boldsymbol{\delta}_u$

      \STATE \textit{(2) Single-step generation (1-NFE)}
      \STATE $\mathbf{z}_{adv} \gets \mathbf{z}_{start}
      - (t_{start} - t_{end}) \cdot \mathbf{u}_{attack}$
      \STATE $\mathbf{x}_{adv} \gets \mathcal{D}(\mathbf{z}_{adv})$

      \STATE \textit{(3) Compute gray-box losses}
      \STATE $\mathcal{F}_{adv} \gets \Phi(\mathbf{x}_{adv})$
      \STATE $\mathcal{L}_{feat} \gets
      \| \mathbf{M} \odot (\mathcal{F}_{adv} - \mathcal{F}_{clean}) \|_2^2$
      \STATE $\mathcal{L}_{attn} \gets
      \| \mathbf{W}_{attn} \odot (\mathcal{F}_{adv} - \mathcal{F}_{clean}) \|_2^2$
      \STATE $\mathcal{L}_{constraint} \gets
      |\sigma(\mathbf{z}_{adv}) - \sigma(\mathbf{z}_{clean})|$
      \STATE $\mathcal{L}_{total} \gets
      - \lambda_f \mathcal{L}_{feat}
      - \lambda_a \mathcal{L}_{attn}
      + \lambda_{c} \mathcal{L}_{constraint}$

      \STATE \textit{(4) Update perturbations}
      \STATE $\boldsymbol{\delta}_z \gets
      \text{Clip}_{\epsilon_z}
      (\boldsymbol{\delta}_z - \eta_z \nabla_{\delta_z}\mathcal{L}_{total})$
      \STATE $\boldsymbol{\delta}_u \gets
      \text{Clip}_{\epsilon_u}
      (\boldsymbol{\delta}_u - \eta_u \nabla_{\delta_u}\mathcal{L}_{total})$
    \ENDFOR
    \STATE \textbf{Final Generation:}
    \STATE Recalculate $\mathbf{z}_{final}$ using optimized $\boldsymbol{\delta}_z, \boldsymbol{\delta}_u$
    \STATE \textbf{return} $\mathbf{x}_{adv} \gets \mathcal{D}(\mathbf{z}_{final})$

  \end{algorithmic}
\end{algorithm}

Remarks on \cref{alg:afm}: As shown in the pseudo-code, the core efficiency stems from the single-step generation (Lines 13-15), which circumvents the costly iterative integration typical of diffusion models. The optimization target is twofold: $\boldsymbol{\delta}_z$ disrupts the semantic layout in the latent space, while $\boldsymbol{\delta}_u$ manipulates the velocity field, steering the generation of the adversarial example. These perturbations are jointly updated (Lines 24-25) by maximizing the disruption to the victim's attention mechanisms while maintaining visual fidelity constraints.

\subsubsection{\textbf{Flow Matching--guided Generative Mechanism}}
We adopt Flow Matching to facilitate efficient one-step adversarial generation directly in the latent space.
By leveraging a neural average velocity field $\nabla \mathbf{u}(\mathbf{z}_{t-1}, r, t)$, we capture the global direction of the probability path.
This approach enables direct state transitions without the need for iterative numerical integration, ensuring both computational efficiency and effective adversarial perturbation.

Formally, the average velocity is defined as the normalized displacement from a
reference latent state $\mathbf{z}_r$ at time $r$ to the current state $\mathbf{z}_t$:
\begin{equation}
\label{eq:mf_def}
\mathbf{z}_t = \mathbf{z}_r + (t - r)\, \mathbf{u}(\mathbf{z}_t, r, t).
\end{equation}

Differentiating \cref{eq:mf_def} with respect to $t$ yields the fundamental identity connecting the instantaneous and average velocities:
\begin{equation}
\label{eq:mf_identity}
\mathbf{u}(\mathbf{z}_t, r, t)
= \mathbf{v}(\mathbf{z}_t, t)
- (t - r)\, \frac{d}{d t}\mathbf{u}(\mathbf{z}_t, r, t).
\end{equation}

\cref{eq:mf_identity} bridges local flow dynamics with global state transitions. Unlike conventional diffusion-based methods that necessitate repeated integration, the Flow Matching identity allows us to directly optimize the average velocity field $\mathbf{u}$. Once learned, this enables one-step generation (1-NFE) at inference time, effectively creating a direct jump from the source noise distribution to the target data distribution along a straight-line trajectory.

Building upon this efficient transport property, we construct adversarial examples by steering the generative trajectory toward adversarial manifolds.
Given a clean sample $\mathbf{x}_{clean}$, we first map it to the latent space via the VAE encoder, $\mathbf{z}_{clean} = \mathcal{E}(\mathbf{x}_{clean})$.
Leveraging the reversibility of Flow Matching, we invert the flow trajectory in a single step to obtain the corresponding noise-level latent anchor:
\begin{equation}
\mathbf{z}_{mid} = \mathcal{S}_{0 \to 1}(\mathbf{z}_{clean}),
\end{equation}
which serves as the initialization point for our attack.

We introduce a dual-perturbation mechanism: a latent perturbation $\boldsymbol{\delta}_z$ that shifts the anchor state, and a velocity perturbation $\boldsymbol{\delta}_u$ that explicitly warps the generative flow. The perturbed start state and velocity field are defined as:
\begin{equation}
\mathbf{z}_{start} = \mathbf{z}_{mid} + \boldsymbol{\delta}_z,
\end{equation}
\begin{equation}
\mathbf{u}_{attack} = \mathcal{S}_\theta(\mathbf{z}_{start}, t_{start}, t_{end}) + \boldsymbol{\delta}_u,
\end{equation}
where $\mathcal{S}_\theta(\cdot)$ is the pre-trained Flow Matching network. 

The adversarial latent code is synthesized via a differentiable single-step linear ODE update:
\begin{equation}
\mathbf{z}_{adv} = \mathbf{z}_{start}
- (t_{start} - t_{end}) \cdot \mathbf{u}_{attack}.
\end{equation}
Following the standard convention, we set $t_{start}=1$ and $t_{end}=0$ to represent the reverse transport from noise to data.
Finally, the adversarial image is reconstructed via the VAE decoder, $\mathbf{x}_{adv} = \mathcal{D}(\mathbf{z}_{adv})$.

\subsubsection{\textbf{Attention-Guided Objective}}
To maximize the impact on driving decisions while maintaining visual imperceptibility, we design a composite objective function.
Note that during the attack, all network parameters are fixed; gradients are used solely to update the perturbations.
The total loss $\mathcal{L}_{total}$ to be minimized is defined as:
\begin{equation}
\mathcal{L}_{total} = \underbrace{-\lambda_{f}\mathcal{L}_{feat} - \lambda_{a}\mathcal{L}_{attn}}_{\text{Maximization (Attack)}} + \underbrace{\lambda_{c}\mathcal{L}_{constraint}}_{\text{Minimization (Regularization)}},
\end{equation}
where $\lambda$ terms are weighting coefficients.

\textit{(i) Road-Focused Feature Loss.}
Driving decisions are primarily governed by road topology (e.g., lanes) rather than background regions (e.g., sky). To incorporate this prior, we construct a spatial mask $\mathbf{M}$ that assigns higher weights ($w_{road}=3.0$) to the lower regions of the feature map. We maximize the weighted feature deviation:
\begin{equation}
\mathcal{L}_{feat} = \frac{1}{|\mathcal{F}|} \sum_{i,j} \mathbf{M}_{i,j} \left\| \mathcal{F}^{(i,j)}_{adv} - \mathcal{F}^{(i,j)}_{clean} \right\|_2^2,
\end{equation}
where $\mathcal{F}$ denotes feature embeddings from the Transformer backbone. This encourages disruption specifically in decision-critical regions.

\textit{(ii) Attention-Weighted Loss.}
We further exploit Transformer-specific inductive biases by leveraging internal attention maps. Specifically, we extract the global token saliency map $\mathbf{A}_{clean}$  
from the Transformer backbone, which quantifies the model's reliance on each spatial token. To target high-saliency regions, we compute the attention-weighted deviation with temperature scaling ($\tau = 4.5$):
\begin{equation}
\mathcal{L}_{attn} = \sum_{k=1}^{N} \text{Softmax} \left( \mathbf{A}^{(k)}_{clean} / \tau \right) \left\| \mathcal{F}^{(k)}_{adv} - \mathcal{F}^{(k)}_{clean} \right\|_2^2,
\end{equation}
where $N$ denotes the number of tokens.

\textit{(iii) Latent Anchor Constraint.}
To ensure the synthesized samples remain visually realistic, we enforce a statistical constraint in the latent space:
\begin{equation}
\mathcal{L}_{constraint} = \left| \sigma(\mathbf{z}_{adv}) - \sigma(\mathbf{z}_{clean}) \right|,
\end{equation}
where $\sigma(\cdot)$ denotes the standard deviation. This prevents distribution collapse and ensures the adversarial latent code remains on the natural image manifold.

\begin{figure*}[h]
    \centering
    \includegraphics[width=1\textwidth]{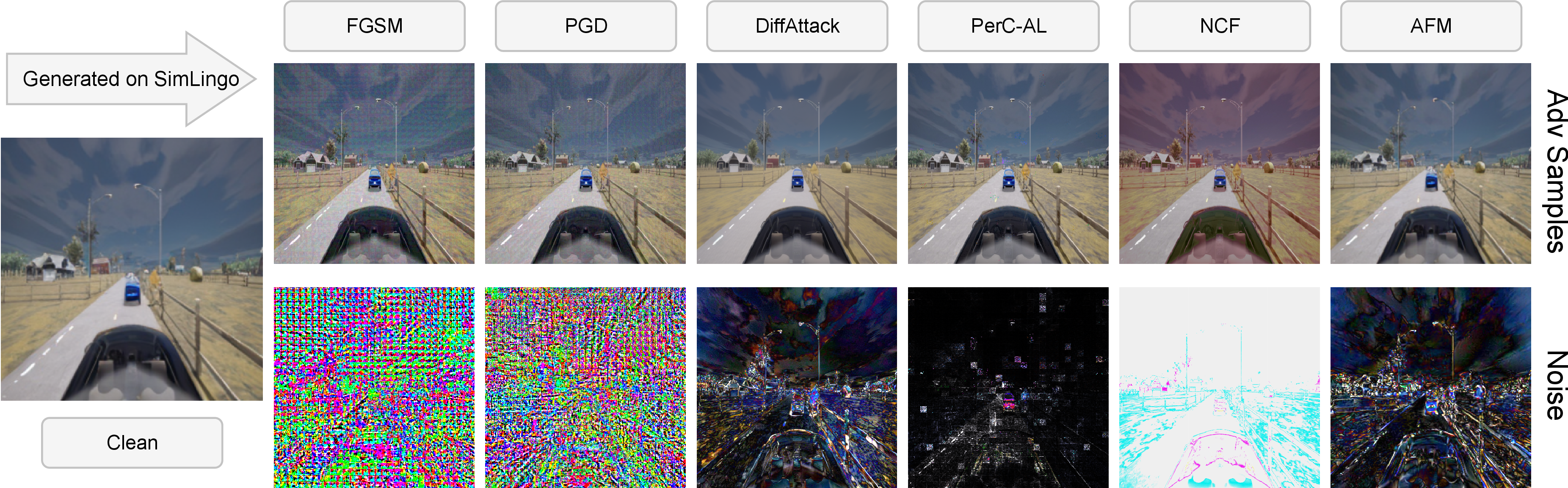}
    \caption{\emph{Qualitative comparison of adversarial examples and corresponding perturbation noise generated from Simlingo.} The top row shows adversarial images generated by baseline methods (FGSM, PGD, DiffAttack, PerC-AL, NCF) and the proposed AFM, while the bottom row visualizes the corresponding perturbations to facilitate visual comparison. Baseline methods often produce visually noticeable high-frequency patterns or localized color distortions, whereas AFM yields smoother perturbations that are more spatially coherent and remain difficult to perceive under normal viewing conditions.}
    \label{fig:qualitative_noise}
    \vspace{-5pt}
\end{figure*}

\section{Experiments}

\subsection{\textbf{Experimental Setup}}
\subsubsection{\textbf{Representative Models}}
To evaluate the universality of our proposed attack across diverse AD agent paradigms, we select two representative models from distinct architectural categories: TransFuser \cite{chitta2022transfuser} for specialized modular architectures and SimLingo \cite{renz2025simlingo} for monolithic VLA architectures.
Despite their structural differences, both models integrate Transformer modules as core components for feature processing. This shared characteristic satisfies the structural prerequisite of our gray-box framework, which specifically targets visual representations within the Transformer backbone. 
We specifically elaborate on their visual encoding and information integration mechanisms to elucidate the vulnerability surfaces exploited by our proposed attack method.

TransFuser employs a specialized modular architecture that processes RGB images and LiDAR BEV inputs via modality-specific encoders, utilizing dual-stream RegNet backbones to extract features at multiple resolutions. Distinct from standard late-fusion approaches, TransFuser incorporates Transformer modules at intermediate layers of the encoders to fuse features from the perspective view and the bird's-eye view. These modules utilize self-attention mechanisms to model global dependencies and capture the geometric correspondence between different modalities throughout the feature extraction process. For this model, we specifically target the visual modality by exploiting these Transformer-based fusion layers, ensuring that perturbations to visual features effectively propagate through the multi-modal integration process and disrupt the global context reasoning required for safe planning.

SimLingo represents the monolithic VLA paradigm, built upon the advanced InternVL2 architecture. It utilizes a robust vision foundation model, specifically InternViT, which employs a dynamic high-resolution strategy to split input images into local tiles and a global thumbnail for comprehensive feature extraction. These multi-scale visual features are then downsampled via pixel unshuffling, projected into the language embedding space via a Multi-Layer Perceptron (MLP), and fed into a Transformer-based Large Language Model (Qwen2) for high-level reasoning and decision-making. Here, we target the ViT component to perturb the initial visual encoding chain. By manipulating the continuous visual tokens before they enter the LLM, we can fundamentally mislead the downstream reasoning process, causing the model to generate unsafe action predictions.

\subsubsection{\textbf{Datasets and Scenario Splits}}
To ensure a rigorous evaluation aligned with the native training distributions of the selected architectures, we utilize their respective official datasets. 
Specifically, we detail the exact number of frames, route distributions, and scenario types for both TransFuser and SimLingo to ensure the reproducibility of our experimental results.

\textit{Dataset for Specialized Modular Architecture (TransFuser).} We use the standard imitation learning dataset provided by Chitta et al \cite{chitta2022transfuser}, comprising 228k frames collected at 2 Hz by a privileged rule-based expert in CARLA. To evaluate robustness across varying traffic complexities, we stratify the dataset into two distinct subsets: Complex Scenarios, characterized by dense traffic at unsignalized intersections, and Common Scenarios, representing standard highway navigation.

\textit{Dataset for VLA Architecture (SimLingo).} We employ the large-scale multimodal dataset from Renz et al \cite{renz2025simlingo}, containing 3.1 million frames collected at 4 Hz via the PDM-lite expert. To evaluate the efficacy of our attack under diverse environmental conditions, we select samples from distinct illumination settings: Daytime Scenarios and Nighttime Scenarios, derived from the simulation's diverse weather presets.

\subsubsection{\textbf{Implementation Details}} 
The dual perturbations $\boldsymbol{\delta}_z$ and $\boldsymbol{\delta}_u$ are optimized using the Adam optimizer for $N=50$ iterations.
We set the learning rates to $\eta_z = 0.05$ and $\eta_u = 0.05$.
To strictly enforce imperceptibility, the perturbation magnitudes are clipped to $\epsilon_z = 0.03$ and $\epsilon_u = 0.03$ within the latent space.
The weighting coefficients for the composite objective function are configured as follows: $\lambda_f = 3.0$ for the road-focused feature loss, $\lambda_a = 4.5$ for the attention-weighted loss, and $\lambda_c = 6.0$ for the latent anchor constraint.
The temperature parameter for attention sharpening is set to $\tau = 4.5$.
For the spatial mask $\mathbf{M}$, we assign a weight of $3.0$ to the road-centric region (defined as the lower $45\%$ of the feature map) and $1.0$ to the background.
All  experiments are implemented in PyTorch. Open-loop experiments were conducted on a single NVIDIA GeForce RTX 4090 GPU, while closed-loop experiments were conducted on a single A800 80G GPU. 

\subsection{\textbf{Baseline Methods}}
\subsubsection{\textbf{Standard Gradient-based Attack Baselines}} 
We include the FGSM \cite{goodfellow2014explaining} and PGD \cite{madry2017towards} as standard gradient-based benchmarks. FGSM applies a one-step perturbation guided by the sign of the input gradient, whereas PGD performs iterative gradient updates projected onto an $\epsilon$-ball constraint.

\subsubsection{\textbf{Generative and Perceptual Attack Baselines}} 
Beyond pixel-level perturbations, we compare against attacks emphasizing semantic fidelity and visual imperceptibility. DiffAttack \cite{chen2024diffusion} serves as a representative generative attack leveraging diffusion models. Unlike conventional pixel-wise manipulation, it operates within the latent space of diffusion models to synthesize adversarial examples while preserving high-level semantic structure. Additionally, we include PerC-AL \cite{zhao2020towards} and NCF \cite{yuan2022natural} as perceptual attack methods. PerC-AL operates in the CIELAB color space, employing an alternating optimization objective to generate perceptually aligned perturbations. Similarly, NCF performs unrestricted color manipulations by adjusting local color distributions based on realistic color sampling, producing natural-looking adversarial examples that subtly influence agent behavior.
Beyond quantitative metrics, \Cref{fig:qualitative_noise} provides a qualitative visualization of adversarial examples and their perturbation patterns, highlighting the distinct characteristics of different attack paradigms in terms of spatial coherence and perceptual visibility.

\begin{figure*}[h]
    \centering
    \includegraphics[width=1\textwidth]{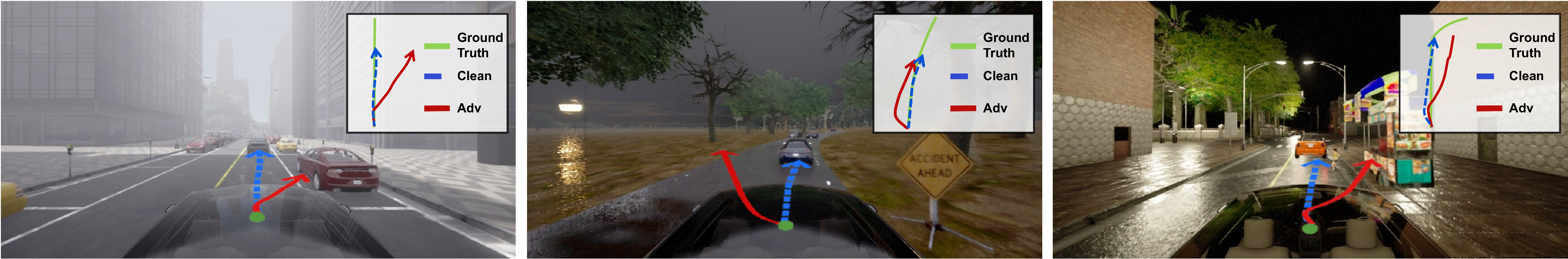}
    \caption{\emph{Qualitative illustration of trajectory deviations induced by AFM on Simlingo.} Clean trajectories (blue) and adversarial trajectories (red) are shown for representative driving scenarios. This visualization highlights the characteristic deviation trends caused by adversarial perturbations, including pronounced lateral offsets that can result in lane departures or collisions, rather than exact waypoint predictions.}
    \label{fig:trajectory_shift}
\end{figure*}

\subsection{\textbf{Evaluation Metrics}}
We evaluate the proposed framework comprehensively through both open-loop prediction analysis and closed-loop dynamic simulations. Specifically, our assessment spans four complementary dimensions: open-loop attack effectiveness, computational efficiency, visual imperceptibility, and system-level closed-loop driving performance. 
Attack effectiveness is measured using Mean Trajectory Shift (SHIFT), which quantifies the displacement of predicted waypoints, and Attack Success Rate (SR), which reflects the frequency of induced safety violations. 
For intuitive understanding, we further provide an illustrative visualization of predicted trajectories under clean and adversarial samples in \Cref{fig:trajectory_shift}, highlighting how visually imperceptible perturbations can translate into substantial deviations in driving behavior. 
Computational efficiency is evaluated via the average adversarial generation time (TIME), indicating the practical feasibility of the attack in real-time constraints. 
To assess visual imperceptibility, we adopt SSIM \cite{wang2004image}, LPIPS \cite{Zhang_2018_CVPR}, and Frechet Inception Distance (FID) \cite{NIPS2017_8a1d6947} to measure structural similarity, perceptual distance, and distributional realism, respectively. 
Finally, closed-loop driving performance is rigorously evaluated using Route Completion (RC) alongside a granular breakdown of physical failure modes to capture the temporal accumulation of adversarial hazards in interactive environments.

\subsubsection{\textbf{Attack Effectiveness}}
The primary goal of the attack is to deviate the autonomous agent from its planned trajectory and induce safety violations. To quantify this, we utilize Mean Trajectory Shift (SHIFT) and Attack Success Rate
(SR). The SHIFT metric measures the average Euclidean distance between the predicted waypoints of the clean sample $\mathbf{x}$ and the adversarial sample $\mathbf{x}'$ over the prediction horizon $T$.
It is formally defined as
\begin{equation}
\text{SHIFT} = \frac{1}{N} \sum_{i=1}^{N} \left( \frac{1}{T} \sum_{t=1}^{T} \left\| \mathcal{M}(\mathbf{x}_i)_t - \mathcal{M}(\mathbf{x}'_i)_t \right\|_2 \right),
\end{equation}
where $N$ denotes the total number of samples, and $\mathcal{M}(\cdot)_t \in \mathbb{R}^2$ represents the model's predicted waypoint at time step $t$. A higher SHIFT value indicates a stronger deviation from the model’s planned trajectory or intended behavior. Complementing this, the SR quantifies the success of the attack. An attack is considered successful if it causes a collision, an off-road event, or a lane deviation exceeding a predefined safety threshold. SR is calculated as the ratio of successful attacks $N_{succ}$ to the total number of attempts $N_{total}$:
\begin{equation}
\text{SR} = \frac{N_{succ}}{N_{total}} \times 100\%.
\end{equation}

\subsubsection{\textbf{Computational Efficiency}}
Beyond effectiveness, the practical applicability of an adversarial attack in dynamic driving scenarios depends on its latency. We evaluate this using Generation Time (TIME). Let $t_{gen}(\mathbf{x}_i)$ denote the wall-clock time consumed to generate the
adversarial perturbation for the $i$-th input sample $\mathbf{x}_i$.
The average generation time is defined as
\begin{equation}
\text{TIME} = \frac{1}{N} \sum_{i=1}^{N} t_{gen}(\mathbf{x}_i),
\end{equation}
where $N$ is the total number of evaluation samples. Low latency is crucial for ensuring that the attack algorithm can operate within the real-time constraints of the end-to-end autonomous driving agent.

\subsubsection{\textbf{Visual Imperceptibility}}
To ensure the generated perturbations remain inconspicuous to human observers and preserve the statistical properties of the original domain, we employ three perceptual metrics: Structural Similarity Index (SSIM), Learned Perceptual Image Patch Similarity (LPIPS), and Fréchet Inception Distance (FID). First, SSIM evaluates the degradation of visual quality by analyzing luminance, contrast, and structural information:
\begin{equation}
\text{SSIM}(x, x') = \frac{(2\mu_x\mu_{x'} + c_1)(2\sigma_{xx'} + c_2)} {(\mu_x^2 + \mu_{x'}^2 + c_1)(\sigma_x^2 + \sigma_{x'}^2 + c_2)},
\end{equation}
where $\mu$ and $\sigma$ denote the mean and variance of image intensities. To capture perceptual differences that align better with human vision, we verify the deep feature distance using LPIPS. Let $\mathcal{F}$ denote a pre-trained feature extraction network. The metric computes the weighted Euclidean distance between the feature maps of the clean image $x$ and the adversarial image $x'$ across $L$ layers:
\begin{equation}
\text{LPIPS}(x, x') = \sum_{l=1}^{L} \frac{1}{H_l W_l} \sum_{h,w} \left\| w_l \odot \left( \hat{\mathbf{y}}^l_{hw} - \hat{\mathbf{y}}'^l_{hw} \right) \right\|_2^2,
\end{equation}
where $\hat{\mathbf{y}}^l$ represents the unit-normalized feature stack from layer $l$. Finally, we utilize FID to measure the distributional distance between the clean images ($\mathcal{D}_x$) and adversarial images ($\mathcal{D}_{x'}$) in the feature space of an Inception-v3 network. Assuming these features follow a multivariate Gaussian distribution, FID is calculated as
\begin{equation}
\begin{aligned}
\text{FID}(\mathcal{D}_x, \mathcal{D}_{x'}) &= \| \mu_x - \mu_{x'} \|_2^2 \\ &\quad + \mathrm{Tr}\!\left(\Sigma_x + \Sigma_{x'} - 2(\Sigma_x \Sigma_{x'})^{1/2} \right).
\end{aligned}
\end{equation}
Lower LPIPS and FID scores imply higher imperceptibility and realism.

\subsubsection{\textbf{Closed-Loop Benchmark}}
To rigorously assess the impact of adversarial attacks in dynamic, interactive environments, we incorporate the standard closed-loop evaluation metrics from the Bench2Drive benchmark. Unlike open-loop metrics that evaluate single-step prediction deviations, these closed-loop metrics capture the temporal accumulation of errors and profile the resulting physical failure modes. 

The primary macro metric we chose is \textbf{Route Completion (RC)}, which measures the percentage of the prescribed route successfully navigated by the ego-vehicle before an episode terminates. To provide a granular analysis of the specific hazards induced by the attacks, we monitor several critical categorical infractions:
\begin{itemize}
    \item \textbf{Off-Road}: The occurrence rate of the ego-vehicle departing from the designated drivable area.
    \item \textbf{Collisions}: The frequency of physical impacts with dynamic objects (e.g., other vehicles, pedestrians) or static environmental obstacles.
    \item \textbf{Route Deviation (Route Dev)}: Instances where the vehicle severely drifts or deviates from the assigned global navigational trajectory.
    \item \textbf{Blocked}: Situations where the agent improperly stalls or remains stationary for an extended duration, failing to make forward progress.
    \item \textbf{Timeouts}: Temporal failures, specifically divided into \textbf{Scenario Timeout} (failing to navigate a specific local traffic scenario within the expected time) and \textbf{Route Timeout} (exceeding the maximum allocated time for the entire route).
\end{itemize}

By jointly analyzing RC alongside these specific failure modes, we can effectively distinguish between low-severity disruptions (e.g., static freezing characterized by high Blocked rates) and high-severity active hijacking (e.g., elevated Off-Road and Route Dev rates), thereby providing a holistic and realistic assessment of attack destructiveness.

\subsection{\textbf{Results}}
\subsubsection{\textbf{Quantitative Comparison of Attack Performance}}
The quantitative performance of different adversarial attack methods on TransFuser and SimLingo across diverse driving scenarios is reported in \Cref{tab:transfuser} and \Cref{tab:simlingo}, respectively.

Across both benchmarks, AFM achieves the strongest visual imperceptibility among all evaluated methods. On TransFuser Common scenarios (\Cref{tab:transfuser}), AFM attains an LPIPS of 0.141 and an SSIM of 0.881, which are the best values across all baselines, while also yielding the lowest FID of 23.180. Similarly, on SimLingo Daytime scenarios (\Cref{tab:simlingo}), AFM achieves the lowest LPIPS of 0.074, the lowest FID of 8.875, and an SSIM of 0.961, outperforming all competing attacks in terms of perceptual quality.

\begin{figure*}[h]
    \centering
    \includegraphics[width=0.7\textwidth]{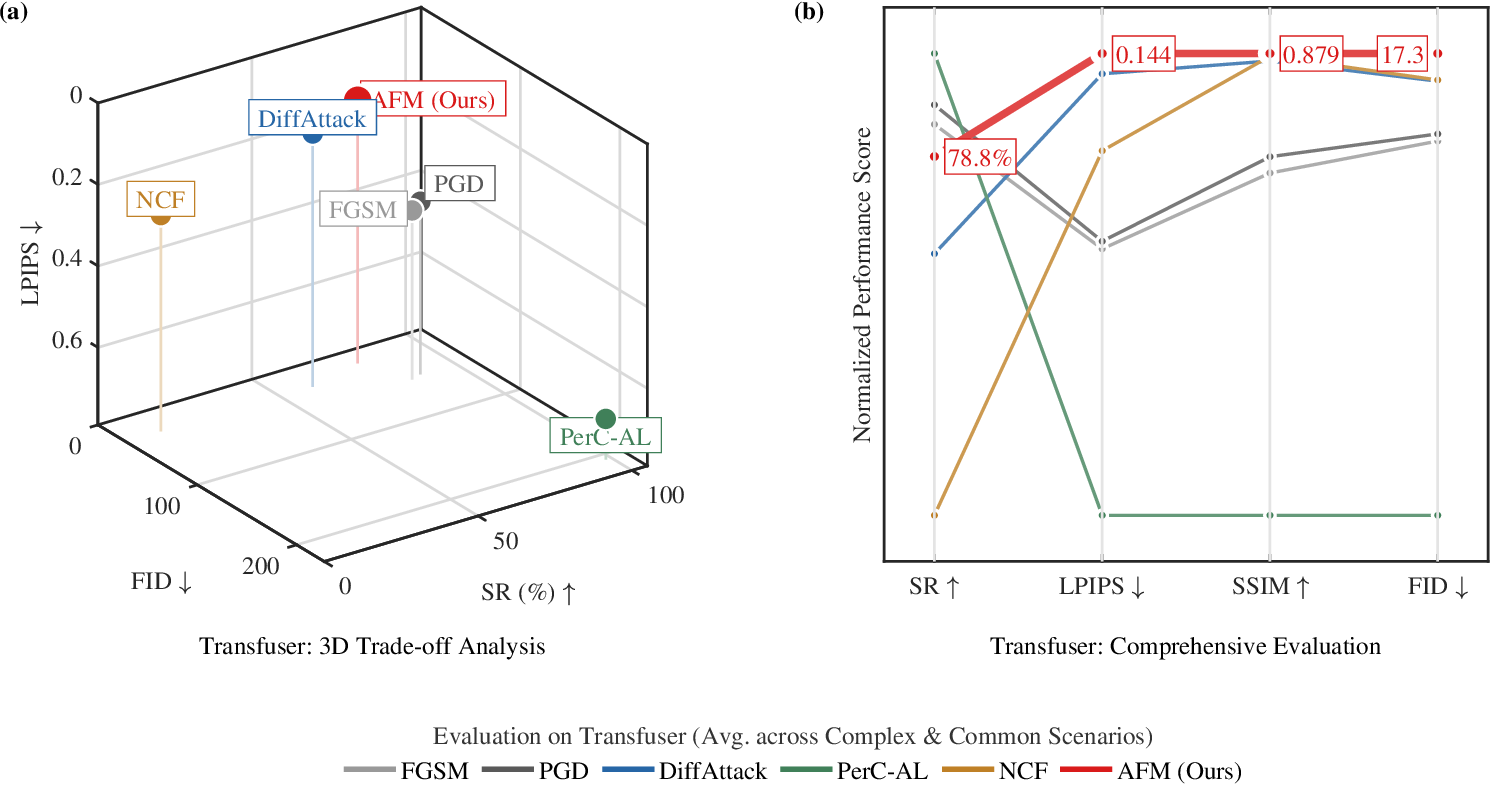}
    \caption{\emph{Quantitative evaluation and trade-off analysis on TransFuser.} (a) 3D Trade-off Analysis: This plot visualizes the equilibrium between attack success rate (SR $\uparrow$) and visual imperceptibility (LPIPS $\downarrow$, FID $\downarrow$) on a specialized modular baseline. AFM occupies the optimal upper-corner of the objective space, achieving the best balance by significantly reducing perceptual distortion while maintaining competitive attack potency.
(b) Aggregated Performance Profiles: A comparison of normalized scores averaged across both complex (H) and common (E) scenarios. The bold red line highlights AFM’s superior performance, particularly in visual imperceptibility (SOTA LPIPS and SSIM), where it outperforms traditional gradient-based methods (FGSM, PGD) and generative baselines (DiffAttack) by a substantial margin.}
    \label{fig:oltransfer}
\end{figure*}

On the TransFuser benchmark (\Cref{tab:transfuser}), PerC-AL achieves the highest attack success rates in both Complex and Common scenarios, reaching 99.28\% and 97.06\%, respectively. However, this performance is accompanied by severe visual degradation, with FID values exceeding 190 and SSIM dropping to approximately 0.2, indicating highly conspicuous perturbations. In contrast, AFM maintains substantially higher image quality, achieving SSIM values above 0.87 in both scenario types while still inducing strong trajectory deviations. Compared with DiffAttack, AFM consistently improves perceptual metrics: for example, in Common scenarios, AFM reduces LPIPS from 0.165 to 0.141 and lowers FID from 36.565 to 23.180, while also increasing SR from 60.12\% to 88.24\%. This optimal balance between attack potency and perceptual quality on the modular baseline is intuitively visualized in the 3D objective space and aggregated profiles in \Cref{fig:oltransfer}.

\begin{figure*}[h]
    \centering
    \includegraphics[width=0.7\textwidth]{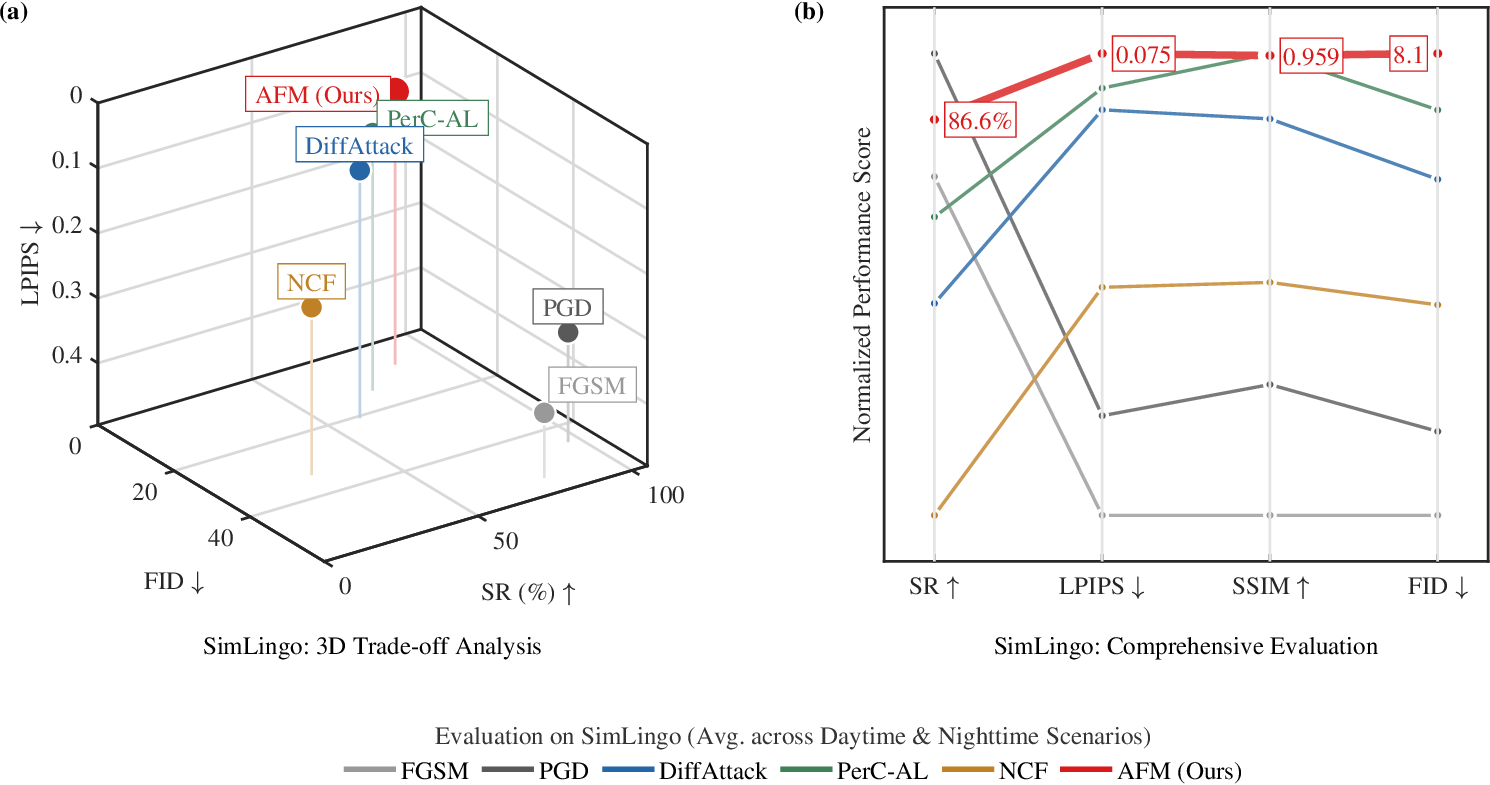}
    \caption{\emph{Quantitative evaluation and trade-off analysis on SimLingo.} (a) 3D Trade-off Analysis: This plot visualizes the equilibrium between attack success rate (SR $\uparrow$) and visual imperceptibility (LPIPS $\downarrow$, FID $\downarrow$) on a Vision-Language-Action (VLA) baseline. AFM consistently achieves the optimal balance, situated at the upper-leftmost corner of the 3D objective space, demonstrating superior performance in generating high-potency adversarial samples with minimal perceptual trace.
    (b) Aggregated Performance Profiles: A multi-dimensional comparison across four key metrics, with scores averaged across Daytime (D) and Nighttime (N) scenarios. The bold red line highlights AFM’s superior performance; it not only secures state-of-the-art results in visual stealthiness (surpassing all baselines in LPIPS and FID) but also maintains high attack effectiveness, highlighting the method's cross-scenario robustness and architectural generalizability.}
    \label{fig:olsimlingo}
\end{figure*}

On SimLingo (\Cref{tab:simlingo}), AFM consistently ranks among the most effective attackers under both Daytime and Nighttime conditions. Under Daytime scenarios, AFM achieves an SR of 87.14\%, outperforming DiffAttack (59.98\%) and PerC-AL (71.42\%), while simultaneously attaining the lowest LPIPS and FID values. A similar trend is observed under Nighttime conditions, where AFM reaches an SR of 85.98\% and yields the lowest FID of 7.332, indicating strong attack effectiveness with minimal perceptual distortion. The corresponding comprehensive trade-off landscape for this VLA-based agent is explicitly illustrated in \Cref{fig:olsimlingo}, demonstrating AFM's capacity to minimize perceptual traces while maintaining high attack effectiveness.

Across both benchmarks, performance degradation is observed under more challenging conditions, specifically Complex scenarios for TransFuser and Nighttime scenarios for SimLingo. Despite this, AFM maintains stable attack success and consistently favorable perceptual metrics, whereas several baselines exhibit either sharp drops in SR or pronounced degradation in visual quality. Finally, in terms of computational efficiency, FGSM and PGD remain the fastest methods due to their simple gradient-based formulations. Among advanced attacks, AFM achieves an average generation time of approximately 6.7 seconds per sample, which is substantially lower than DiffAttack (around 10 seconds) and PerC-AL (around 11 seconds), providing a more balanced trade-off between attack effectiveness, perceptual quality, and computational cost.

\subsubsection{\textbf{Computational Efficiency and Practical Deployment Strategies}}
While generative methods naturally incur higher latency than simple gradient-based additive noise, AFM's 1-NFE formulation makes it significantly more efficient than state-of-the-art diffusion-based attacks. 

To address the computational budget of generative adversarial attacks, we report the end-to-end latency of AFM. As shown in Table 1 and Table 2, generating one adversarial frame via AFM on a single NVIDIA RTX 4090 GPU takes approximately 6.7 seconds. While this generation latency is naturally higher than that of simple, non-stealthy gradient-based methods (e.g., FGSM or PGD, which take 0.1 to 1.8 seconds but suffer from severe visual artifacts), AFM demonstrates a significant computational advantage within the paradigm of high-fidelity generative attacks. State-of-the-art diffusion-based attacks (e.g., DiffAttack) typically require multi-step iterative denoising, accumulating a substantial latency of 10 to 11 seconds per image. By leveraging the straight-path property of Flow Matching, AFM achieves the same or better stealthiness in a single integration step (1-NFE), making it nearly 40\% faster than diffusion baselines and substantially reducing the computational overhead.

Furthermore, while a 6.7-second latency currently precludes dense, real-time per-frame generation at 30 Hz, we emphasize that continuous video-stream poisoning is not a prerequisite for successful AD attacks. To bypass the real-time inference bottleneck, we outline two highly realistic deployment strategies for AFM in the physical world:
\begin{itemize}
    \item {\textbf{Intermittent Digital Injection:} VLA-based AD agents rely heavily on temporal reasoning. Injecting a sparsely generated AFM adversarial frame at critical timestamps (e.g., just before approaching an intersection) is sufficient to disrupt the temporal context and derail the planned trajectory, eliminating the need for strict real-time, per-frame generation.}
    \item {\textbf{Offline Physical Attack Proxies:} The generated adversarial perturbations can be utilized as offline priors. For instance, the velocity-field manipulated noise can be printed as physical adversarial patches and overlaid on static real-world objects, such as traffic lights or roadside billboards. In such physical deployment scenarios, the online generative latency is completely bypassed, making our SOTA visual imperceptibility the decisive factor for a successful, undetected attack.}
\end{itemize}

\begin{table*}[t]
\centering
\caption{Performance Comparison of Attack Methods on Transfuser across Complex(H) and Common(E) Scenarios}
\label{tab:transfuser}
\resizebox{0.9\textwidth}{!}{%
\begin{small} 
\setlength{\tabcolsep}{6pt} 
\begin{tabular}{lcccccccccccc}
\toprule
\multirow{2}{*}{\textbf{Method}} 
 & \multicolumn{2}{c}{\textbf{SHIFT (m)} $\uparrow$} & \multicolumn{2}{c}{\textbf{SR (\%)} $\uparrow$} & \multicolumn{2}{c}{\textbf{LPIPS} $\downarrow$} & \multicolumn{2}{c}{\textbf{SSIM} $\uparrow$} & \multicolumn{2}{c}{\textbf{FID} $\downarrow$} & \multicolumn{2}{c}{\textbf{TIME (s)}} \\
\cmidrule(lr){2-3} \cmidrule(lr){4-5} \cmidrule(lr){6-7} \cmidrule(lr){8-9} \cmidrule(lr){10-11} \cmidrule(lr){12-13}
 & \textbf{H} & \textbf{E} & \textbf{H} & \textbf{E} & \textbf{H} & \textbf{E} & \textbf{H} & \textbf{E} & \textbf{H} & \textbf{E} & \textbf{H} & \textbf{E} \\
\midrule
FGSM       & 4.188 & 4.889 & 79.99 & 89.71 & 0.379 & 0.370 & 0.697 & 0.708 & 42.401 & 64.187 & 6.134 & \textbf{0.140} \\
PGD        & \underline{5.792} & \underline{5.997} & \underline{84.98} & \underline{92.03} & 0.372 & 0.359 & 0.722 & 0.732 & 42.271 & 58.425 & \textbf{1.208} & \underline{1.230} \\
DiffAttack & 1.515 & 2.465 & 60.91 & 60.12 & \underline{0.170} & \underline{0.165} & 0.865 & 0.871 & \underline{20.794} & 36.565 & 10.616 & 10.606 \\
PerC-AL    & \textbf{6.741} & \textbf{6.809} & \textbf{99.28} & \textbf{97.06} & 0.685 & 0.694 & 0.202 & 0.196 & 190.116 & 223.980 & 10.979 & 11.153 \\
NCF        & 0.261 & 0.517 & 7.32  & 15.34 & 0.262 & 0.256 & \underline{0.871} & \underline{0.879} & 22.177 & \underline{34.223} & \underline{3.075} & 3.007 \\
\midrule
\textbf{AFM (Ours)} & 3.709 & 4.932 & 69.35 & 88.24 & \textbf{0.147} & \textbf{0.141} & \textbf{0.876} & \textbf{0.881} & \textbf{11.460} & \textbf{23.180} & 6.652 & 6.749 \\
\bottomrule
\end{tabular}
\end{small}
}
\end{table*}

\begin{table*}[t]
\centering
\caption{Performance Comparison of Attack Methods on Simlingo across Daytime(D) and Nighttime(N) Scenarios}
\label{tab:simlingo}
\resizebox{0.9\textwidth}{!}{%
\begin{small}
\setlength{\tabcolsep}{6pt}
\begin{tabular}{lcccccccccccc}
\toprule
\multirow{2}{*}{\textbf{Method}} 
 & \multicolumn{2}{c}{\textbf{SHIFT (m)} $\uparrow$} & \multicolumn{2}{c}{\textbf{SR (\%)} $\uparrow$} & \multicolumn{2}{c}{\textbf{LPIPS} $\downarrow$} & \multicolumn{2}{c}{\textbf{SSIM} $\uparrow$} & \multicolumn{2}{c}{\textbf{FID} $\downarrow$} & \multicolumn{2}{c}{\textbf{TIME (s)}} \\
\cmidrule(lr){2-3} \cmidrule(lr){4-5} \cmidrule(lr){6-7} \cmidrule(lr){8-9} \cmidrule(lr){10-11} \cmidrule(lr){12-13}
 & \textbf{D} & \textbf{N} & \textbf{D} & \textbf{N} & \textbf{D} & \textbf{N} & \textbf{D} & \textbf{N} & \textbf{D} & \textbf{N} & \textbf{D} & \textbf{N} \\
\midrule
FGSM       & \underline{2.638} & \underline{2.414} & 79.48 & 76.87 & 0.407 & 0.382 & 0.745 & 0.723 & 59.370 & 48.653 & \textbf{0.474} & \textbf{0.479} \\
PGD        & \textbf{6.787} & \textbf{6.236} & \textbf{95.48} & \textbf{97.03} & 0.346 & 0.306 & 0.795 & 0.800 & 50.008 & 41.331 & \underline{1.789} & \underline{1.806} \\
DiffAttack & 1.703 & 1.614 & 59.98 & 59.16 & 0.110 & 0.117 & 0.939 & 0.916 & 19.995 & 21.224 & 9.980 & 9.982 \\
PerC-AL    & 2.209 & 2.379 & 71.42 & 73.10 & \underline{0.089} & \underline{0.109} & \textbf{0.973} & \underline{0.947} & \underline{11.008} & \underline{16.416} & 10.830 & 10.835 \\
NCF        & 0.977 & 0.841 & 32.27 & 24.65 & 0.219 & 0.254 & 0.917 & 0.779 & 34.317 & 31.868 & 4.536 & 4.590 \\
\midrule
\textbf{AFM (Ours)} & 3.360 & 3.039 & \underline{87.14} & \underline{85.98} & \textbf{0.074} & \textbf{0.075} & \underline{0.961} & \textbf{0.957} & \textbf{8.875} & \textbf{7.332} & 6.814 & 6.849 \\
\bottomrule
\end{tabular}
\end{small}
}
\end{table*}

\begin{figure}[h]
    \centering
    \hspace*{-0.5cm}
    \includegraphics[width=0.5\textwidth]{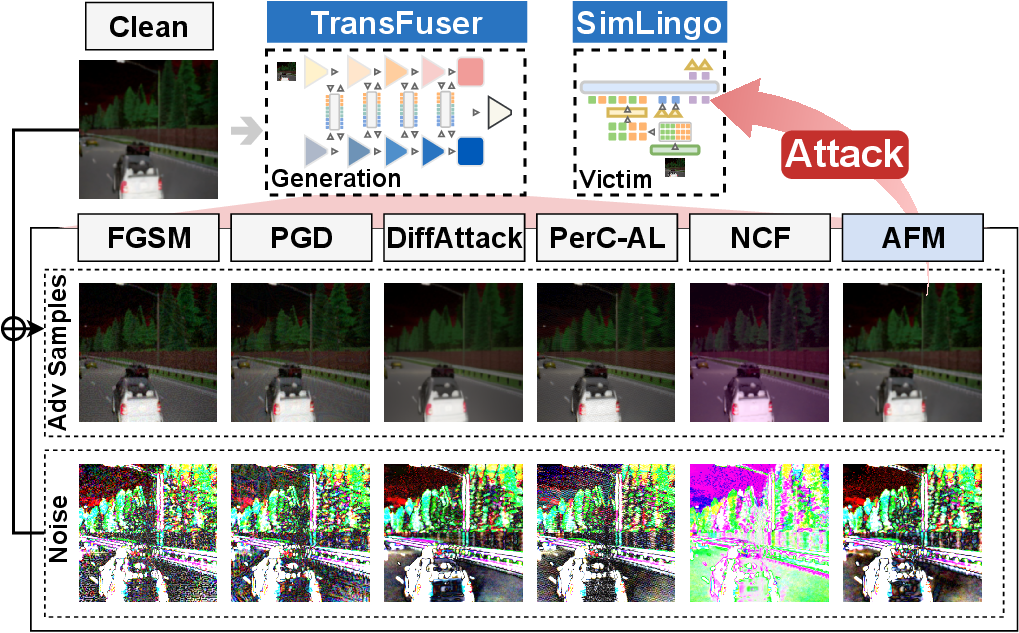}
    \caption{\emph{Transfer attack framework and qualitative results.} Top: The cross-model attack pipeline. Bottom: Visual comparison demonstrating AFM's superior imperceptibility against baselines.}
    \label{fig:transfer_qualitative}
    \vspace{-5pt}
\end{figure}

\begin{table}[h]
\vspace{10pt}
\centering
\caption{Cross-model transferability analysis between SimLingo (SL) and TransFuser (TF).}
\label{tab:cross_model_transfer}
\resizebox{0.5\textwidth}{!}{%
\begin{small}
\setlength{\tabcolsep}{3pt} 
\begin{tabular}{lcccccccc}
\toprule
\multirow{2}{*}{\textbf{Method}} & \multicolumn{2}{c}{\textbf{SHIFT (m)} $\uparrow$} & \multicolumn{2}{c}{\textbf{SR (\%)} $\uparrow$} & \multicolumn{2}{c}{\textbf{LPIPS} $\downarrow$} & \multicolumn{2}{c}{\textbf{SSIM} $\uparrow$} \\
\cmidrule(lr){2-3} \cmidrule(lr){4-5} \cmidrule(lr){6-7} \cmidrule(lr){8-9}
 & \textbf{SL$\to$TF} & \textbf{TF$\to$SL} & \textbf{SL$\to$TF} & \textbf{TF$\to$SL} & \textbf{SL$\to$TF} & \textbf{TF$\to$SL} & \textbf{SL$\to$TF} & \textbf{TF$\to$SL} \\
\midrule
FGSM       & 0.331 & 1.268 & 8.40  & 47.21 & 0.487 & 0.346 & 0.725 & 0.737 \\
PGD        & 0.316 & \textbf{1.337} & 8.40  & \textbf{50.71} & 0.476 & 0.291 & 0.763 & 0.812 \\
DiffAttack & \underline{0.474} & \underline{1.270} & \underline{12.61} & \underline{48.36} & \underline{0.114} & \underline{0.170} & \underline{0.980} & 0.859 \\
PerC-AL    & 0.165 & 1.207 & 3.36  & 45.79 & 0.124 & 0.357 & 0.960 & 0.776 \\
NCF        & 0.390 & 1.206 & 8.61  & 45.00 & 0.293 & 0.264 & 0.894 & \underline{0.868} \\
\midrule
\textbf{AFM (Ours)} & \textbf{0.506} & 1.192 & \textbf{12.82} & 46.68 & \textbf{0.022} & \textbf{0.148} & \textbf{0.995} & \textbf{0.871} \\
\bottomrule
\end{tabular}
\end{small}
}
\end{table}

\begin{figure*}[h]
    \centering
    \includegraphics[width=0.9\textwidth]{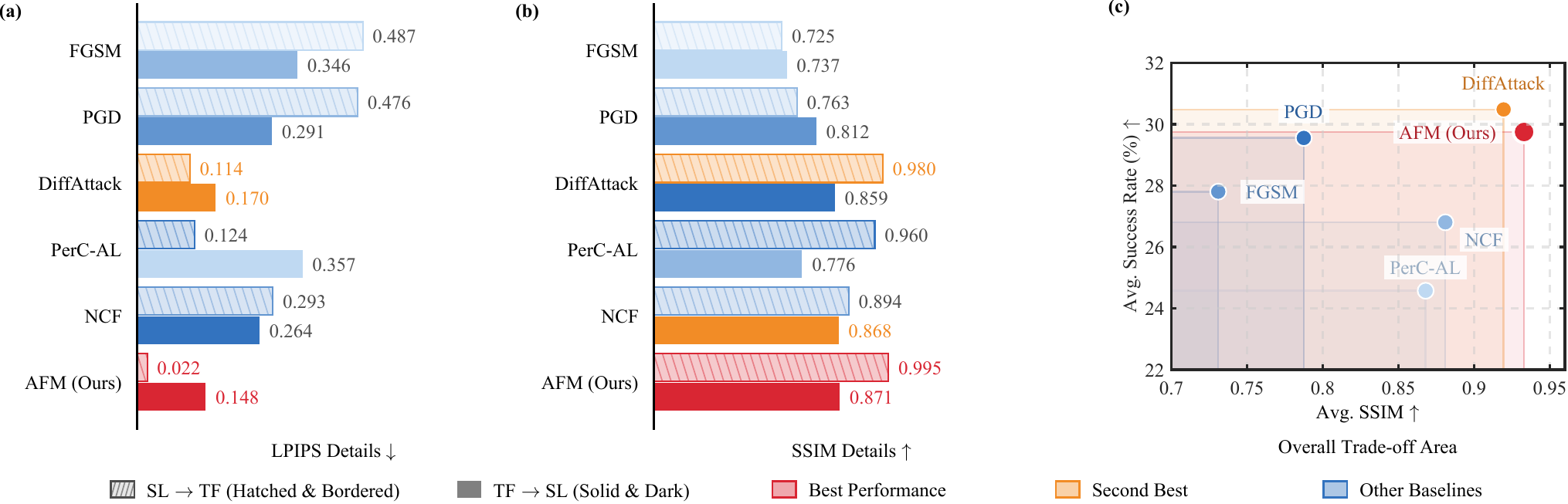}
    \caption{\emph{Cross-model transferability analysis between SimLingo (SL) and TransFuser (TF).} The evaluation covers both transfer directions: generating on the VLA agent to attack the modular agent (SL $\to$ TF, hatched and bordered bars) and the reverse direction (TF $\to$ SL, dark solid bars). Performance is assessed across 
    visual imperceptibility ((a) LPIPS and (b) SSIM) and attack effectiveness (c) Overall Trade-off Area between Avg.SSIM and Avg.Success Rate.
    Bar colors denote performance rankings: red represents the best result (Rank 1), orange represents the runner-up (Rank 2), and blue indicates remaining baselines. AFM consistently maintains highly competitive transfer attack effectiveness while establishing state-of-the-art visual stealthiness across both architectural paradigms.}
    \label{fig:transfer}
\end{figure*}

\subsubsection{\textbf{Cross-Model Transferability Analysis}}
To evaluate cross-model transferability under a strict gray-box setting, we generate adversarial examples on a source architecture and evaluate them on a target architecture without access to the target model’s gradients. To illustrate this process and compare the visual imperceptibility of different attacks, we first present the transfer framework in \Cref{fig:transfer_qualitative}. The comprehensive quantitative evaluation is then summarized in \Cref{tab:cross_model_transfer} and \Cref{fig:transfer} for both transfer directions: transferring attacks from SimLingo to TransFuser (SL $\rightarrow$ TF) and from TransFuser to SimLingo (TF $\rightarrow$ SL).

\emph{SimLingo $\rightarrow$ TransFuser (SL $\rightarrow$ TF).}
As shown in \Cref{tab:cross_model_transfer}, the evaluation metrics result of transfer attacks against the modular TransFuser model using adversarial examples generated on the VLA-based SimLingo. AFM achieves the strongest overall transfer effectiveness, attaining the highest SR and SHIFT among all methods, with SR reaching 12.82\% and SHIFT increasing to 0.506~m. In comparison, PGD yields lower transfer performance with an SR of 8.40\% and a SHIFT of 0.316~m, while FGSM shows a similarly limited SR of 8.40\% despite a slightly higher SHIFT of 0.331~m. Among generative and perceptual baselines under the transfer setting, DiffAttack attains an SR of 12.61\% and a SHIFT of 0.474~m, yet it is consistently outperformed by AFM on both effectiveness metrics. 
As shown by the lighter outlined bars in \Cref{fig:transfer}, AFM achieves the best perceptual quality in this transfer setting, attaining the lowest LPIPS of 0.022 (\Cref{fig:transfer}(a)) and the highest SSIM of 0.995 (\Cref{fig:transfer}(b)). By contrast, DiffAttack and NCF exhibit noticeably higher perceptual distortion, while FGSM and PGD incur severe visual degradation with LPIPS values exceeding 0.470 and SSIM dropping below 0.770, indicating poor visual imperceptibility after transfer.

\emph{TransFuser $\rightarrow$ SimLingo (TF $\rightarrow$ SL).}
We further evaluate the reverse transfer direction, indicated by \Cref{tab:cross_model_transfer}. In this setting, PGD attains the highest SR of 50.71\% and the largest SHIFT of 1.337~m, followed by DiffAttack with an SR of 48.36\% and a SHIFT of 1.270~m. AFM achieves an SR of 46.68\% with a SHIFT of 1.192~m, remaining highly competitive and close to the strongest baselines in terms of transfer effectiveness. Notably, as shown by the solid outlined bars in \Cref{fig:transfer},  AFM provides the strongest visual imperceptibility in this direction, achieving the lowest LPIPS of 0.148 and the highest SSIM of 0.871 among all methods. In contrast, PGD and FGSM exhibit substantially higher perceptual distortion, with LPIPS values of 0.291 and 0.346, respectively, and lower SSIM scores, making them less suitable when imperceptibility is required.

Overall, \Cref{fig:transfer}(c) demonstrates that AFM consistently achieves a strong balance between transfer attack effectiveness and visual imperceptibility. Crucially, these results highlight the formidable potential of AFM as a practical gray-box adversary close to black-box utility. By targeting the intrinsic structural vulnerabilities of attention mechanisms, our findings verify that as long as the victim agent incorporates a Transformer backbone, irrespective of its specific modular or monolithic paradigm, it remains highly susceptible to AFM. This capability allows us to launch highly effective yet visually imperceptible attacks against diverse AD agents, exposing a fundamental shared vulnerability common to these Transformer-based models.

\begin{table*}[t]
\centering
\caption{Comprehensive Comparison of Adversarial Attacks on SimLingo under Closed-Loop Simulations}
\label{tab:comprehensive_attack_comparison}
\resizebox{\textwidth}{!}{%
\begin{tabular}{lccccccccc}
\toprule
\textbf{Method} & \textbf{RC (\%)} $\downarrow$ & \textbf{Off-Road (\%)} $\uparrow$ & \textbf{Collisions} $\uparrow$ & \textbf{Route Dev (\%)} $\uparrow$ & \textbf{Blocked (\%)} $\uparrow$ & \textbf{Sce. Timeout (\%)} $\uparrow$ & \textbf{Route Timeout (\%)} $\uparrow$ & \textbf{LPIPS} $\downarrow$ & \textbf{SSIM} $\uparrow$ \\
\midrule
Clean (Unattacked) & 100.00 & 0.00 & 0.0 & 0 & 0 & 0 & 0 & - & - \\
\midrule
FGSM        & 42.33 & 14.97 & \textbf{2.7} & \underline{10} & \underline{70} & \underline{40} & 0  & 0.594 & 0.511 \\
PGD         & 23.12 & 16.26 & \underline{2.4} & \textbf{20} & \underline{70} & \textbf{50} & 10 & 0.525 & 0.555 \\
DiffAttack  & \underline{5.14}  & 18.63 & 0.6 & 0  & \underline{70} & 20 & \textbf{30} & \underline{0.117} & 0.915 \\
PerC-AL     & 10.63 & \underline{26.08} & 1.4 & \textbf{20} & 50 & \underline{40} & \textbf{30} & 0.237 & \underline{0.922} \\
NCF         & \textbf{0.00}  & 0.00  & 0.0 & 0  & \textbf{100} & 10 & 10 & 0.184 & 0.912 \\
\midrule
\textbf{AFM (Ours)} & 17.29 & \textbf{40.06} & 1.5 & \textbf{20} & 60 & \textbf{50} & \underline{20} & \textbf{0.075} & \textbf{0.956} \\
\bottomrule
\end{tabular}%
}

\vspace{5pt} 
\begin{minipage}{\textwidth}
\footnotesize
\textit{Note:} We report the average performance across all 10 routes of the Bench2Drive benchmark. For categorical test results (Route Deviation (Route Dev), Blocked, Scenario Timeout (Sce. Timeout), Route Timeout), values represent the occurrence failure rate (\%). Arrows indicate the goal of an effective adversarial attack: lower Route Completion (RC) ($\downarrow$) and higher infraction rates ($\uparrow$) imply a more destructive attack, while lower Learned Perceptual Image Patch Similarity (LPIPS) ($\downarrow$) and higher Structural Similarity Index (SSIM) ($\uparrow$) indicate better visual imperceptibility.
\end{minipage}
\end{table*}

\begin{figure*}[t]
    \centering
    \includegraphics[width=0.9\textwidth]{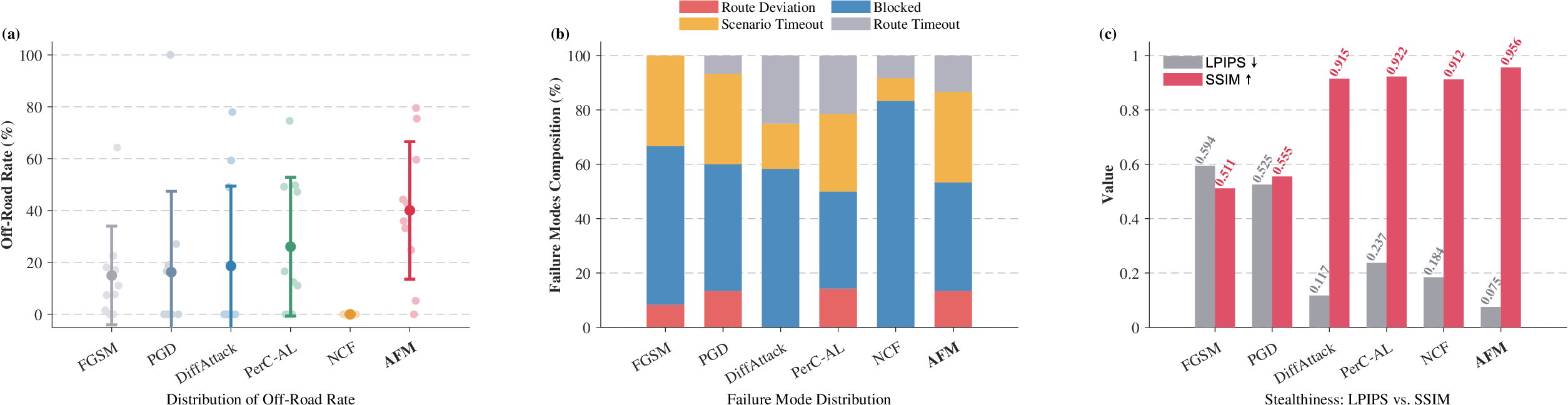}
    \caption{\emph{Statistical analysis of closed-loop driving performance and failure modes on Bench2Drive.} (a) Distribution of Off-Road rates across 10 representative routes. The scatter points denote individual route results, while the solid dots and vertical whiskers represent the mean and standard deviation, respectively. (b) Compositional breakdown of failure modes. To facilitate a fair comparison of attack behaviors, the absolute counts of each failure type are row-normalized to 100\% for each method. While baseline attacks like NCF achieve low RC primarily through "static freezing" (dominated by high Blocked rates), the proposed AFM exhibits superior destructive diversity by actively inducing hazardous maneuvers, characterized by significant timeout rates and route deviations. (c) Stealthiness comparison using LPIPS ($\downarrow$) and SSIM ($\uparrow$) metrics. AFM achieves state-of-the-art visual imperceptibility, significantly outperforming all baseline methods by recording the lowest LPIPS (0.075) and the highest SSIM (0.956).}
    \label{fig:close_loop}
\end{figure*}

\begin{figure*}[h]
    \centering
    \includegraphics[width=0.9\textwidth]{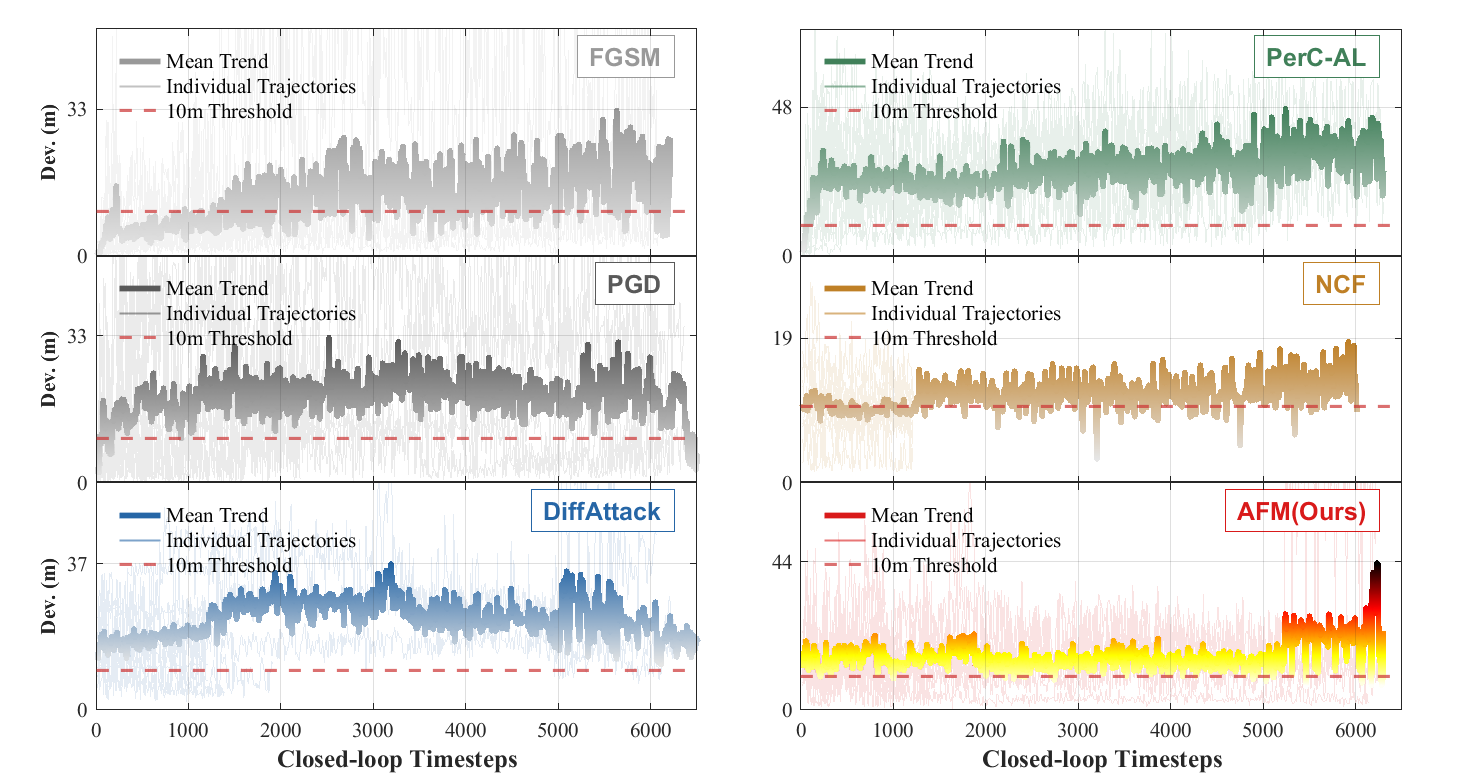}
    \caption{\emph{Temporal evolution of trajectory deviation under various adversarial attacks in closed-loop simulations.} The x-axis represents the elapsed closed-loop timesteps, while the y-axis measures the physical deviation (in meters) from the intended reference route. The red dashed line denotes a critical 10m safety threshold; exceeding this limit indicates severe kinematic infractions such as lane departures or off-road events. While generative baselines like NCF and DiffAttack exhibit bounded deviations or early termination indicative of "static freezing," the proposed AFM demonstrates a continuous, sustained accumulation of compounding errors. This persistent temporal disruption actively hijacks the continuous control policy, ultimately driving the vehicle's trajectory significantly beyond the safety threshold and resulting in catastrophic physical failures.}
    \label{fig:temp_dyna}
\end{figure*}

\subsubsection{\textbf{Closed-Loop Evaluation and Temporal Dynamics}}
To rigorously evaluate the practical hazards of adversarial attacks in dynamic, interactive environments, we conduct comprehensive closed-loop evaluations using the SimLingo agent on the Bench2Drive benchmark. Unlike open-loop evaluations, closed-loop driving is a time-coupled system where agents' inputs heavily depend on previous control actions. By employing a temporal persistence strategy which injecting attacks once every 10 frames, we demonstrate that intermittent strikes are sufficient to trigger a cascade of compounding errors, ultimately leading to catastrophic system failures.

As demonstrated by the macro results in \Cref{tab:comprehensive_attack_comparison}, AFM successfully reconciles the conventional conflict between attack efficacy and visual fidelity. On one hand, achieving meaningful disruption with gradient-based methods (FGSM, PGD) and perceptual baselines (PerC-AL) inevitably introduces noticeable visual artifacts (LPIPS $\ge$ 0.237). AFM, by contrast, establishes a new state-of-the-art in imperceptibility (LPIPS: 0.075, SSIM: 0.956). On the other hand, this extraordinary visual fidelity does not compromise its destructive potential; AFM consistently matches or exceeds the kinematic failure rates of these perceptible baselines.

To deeper understand these compounding failure dynamics, we must look beyond single-dimensional macro metrics like Route Completion (RC), which can be misleading if analyzed in isolation. A granular analysis of the specific metrics in \Cref{tab:comprehensive_attack_comparison}, complemented by the failure mode distributions in \Cref{fig:close_loop} (a) and (b), visually reinforces a fundamental distinction between "static freezing" (where agents primarily fail due to high Blocked rates, as seen in NCF) and "active hijacking" (where the agent is forced into hazardous maneuvers, evidenced by the severe Off-Road infractions induced by AFM).

Generative baselines such as NCF and DiffAttack seemingly exhibit strong attack performance with extremely low average RC (0.00\% and 5.14\%, respectively) and minimal variance. However, a comparative look at their infraction profiles reveals this stems from a trivial vulnerability. As visualized by the temporal evolution of trajectory deviations in \Cref{fig:temp_dyna}, both NCF and DiffAttack fail to induce escalating physical hazards. Their deviation trajectories remain tightly bounded or prematurely terminate, physically confirming a "static freezing" effect. For NCF, the overwhelming 100\% Blocked rate mathematically guarantees a 0.00\% Off-Road rate and 0.0 Collisions; the ego-vehicle simply stalls upon initialization. Similarly, DiffAttack's failure composition is heavily dominated by Blocked events (70\%), resulting in a minimal Collision count (0.6) and zero Route Deviation (Route Dev). These methods paralyze the agent into inaction rather than compromising its driving capability. While they artificially deflate overall RC scores, their over-reliance on static immobilization renders them a low-severity threat in dynamic, real-world traffic where active misdirection is far more dangerous.

In stark contrast, AFM actively derails the continuous control policy to force high-risk maneuvers. By allowing the vehicle to navigate a more substantial portion of the route with more significant route deviations before complete failure, evidenced by the higher average RC of 17.29\% and its broader distribution in \Cref{fig:close_loop} (a), the attack successfully accumulates latent temporal disruptions without attenuation. This compounding error dynamic is explicitly captured in \Cref{fig:temp_dyna}. Unlike the bounded trajectories of generative baselines, AFM demonstrates a continuous, sustained accumulation of deviation over closed-loop timesteps. This persistent temporal disruption actively hijacks the continuous control policy, ultimately driving the vehicle's trajectory sharply beyond the 10m safety threshold to trigger catastrophic physical failures. Specifically, AFM achieves the highest average Off-Road rate of 40.06\%, severely outpacing both DiffAttack (18.63\%) and PerC-AL (26.08\%). Furthermore, as depicted by the diverse failure mode composition in \Cref{fig:close_loop}(b), AFM’s dual-perturbation mechanism effectively hijacks the Transformer's temporal reasoning, translating intermittent spatial manipulations into severe Route Dev (20\% failure rate) and substantial physical Collisions (1.5 on average). 

Finally, \cref{tab:comprehensive_attack_comparison} underscores the unparalleled imperceptibility of AFM amidst this high-intensity, active disruption. Traditional gradient-based attacks (FGSM, PGD) suffer from catastrophic perceptual degradation, yielding average LPIPS scores exceeding 0.525 and SSIM values dropping below 0.560, rendering them easily detectable by standard input-purification defenses. Even advanced perceptual methods like PerC-AL struggle to maintain structural integrity (LPIPS: 0.237). AFM, conversely, maintains state-of-the-art visual stealthiness across all tested scenarios, securing the lowest LPIPS (0.075) and the highest SSIM (0.956). This evidence confirms that synergistically optimizing the neural velocity field and latent space can generate visually imperceptible perturbations that actively mislead predictive planning, representing a far more formidable threat to the long-term stability of AD agents.

\subsection{\textbf{Ablation Studies}}
\label{appendix:ablation_studies}
\begin{figure*}[t] 
  \includegraphics[width=1.0\textwidth]{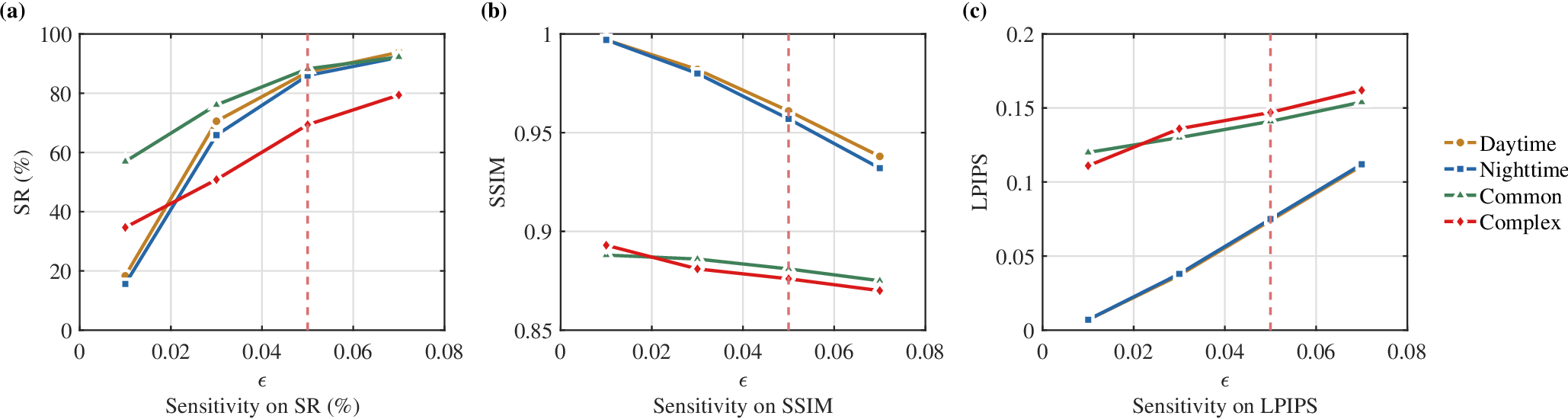}
  \caption{\emph{Ablation study on the perturbation budget $\epsilon$ for AFM.} Increasing $\epsilon$ improves attack effectiveness, reflected by higher SR, while gradually degrading visual fidelity, as indicated by lower SSIM and higher LPIPS. }
  \label{fig:ablation}
\end{figure*}

To better understand the impact of key hyperparameters on the proposed AFM, we conduct ablation studies to examine the trade-off between attack effectiveness and visual imperceptibility. The perturbation budget $\epsilon$ governs the maximum magnitude of adversarial modifications and plays a central role in balancing attack success and perceptual quality. As shown in \Cref{fig:ablation}(a), increasing $\epsilon$ consistently improves the attack success rate (SR), indicating that larger perturbation budgets enable more pronounced trajectory deviations and thus enhance attack effectiveness. However, this improvement comes at the cost of reduced visual fidelity. As illustrated in \Cref{fig:ablation}(b) and \Cref{fig:ablation}(c), larger values of $\epsilon$ lead to lower SSIM and higher LPIPS, reflecting an increasing perceptual discrepancy between clean and adversarial samples. These results reveal an inherent trade-off between attack strength and visual imperceptibility: while stronger perturbations improve attack success, they inevitably introduce more noticeable visual artifacts. Importantly, the selected default perturbation budget used throughout our experiments corresponds to a balanced operating point, achieving strong attack effectiveness while preserving high perceptual quality. Further increasing $\epsilon$ beyond this point yields only marginal gains in attack success but results in a noticeably larger degradation in visual quality.

\section{Conclusion}
In this paper, we presented AFM, a novel gray-box attack framework designed to expose the structural vulnerabilities inherent to the Transformer backbones of AD agents. 
Extensive experiments demonstrate that AFM achieves a superior trade-off between attack effectiveness and imperceptibility. Specifically, it substantially degrades the performance of VLA and modular AD agents across various scenarios while maintaining state-of-the-art visual imperceptibility.
Notably, our transferability experiments have shown that AFM enables effective cross-model attacks solely based on the presence of a Transformer-based module, thereby obviating the need for specific model parameters, architectural details, or gradients, which closely approximates a black-box attack setting.

While our findings are encouraging, the proposed method still has certain limitations. 
First, while more pragmatic, there remains a performance gap between our gray-box approach and full-knowledge white-box attacks. 
Second, our current evaluations are confined to the digital environment, and the proposed attack has yet to be implemented in the physical world. 
To address these constraints, future research will involve a deeper exploration of underlying adversarial mechanisms to enhance attack effectiveness, aiming for white-box potency without compromising visual imperceptibility. Furthermore, we will focus on migrating these techniques to the physical domain, conducting a comprehensive series of validations to ensure robustness against real-world conditions and hardware constraints.
\bibliography{reference}
%

\bibliographystyle{IEEEtran}

\begin{IEEEbiography}[{\includegraphics[width=1in,height=1.25in,clip,keepaspectratio]{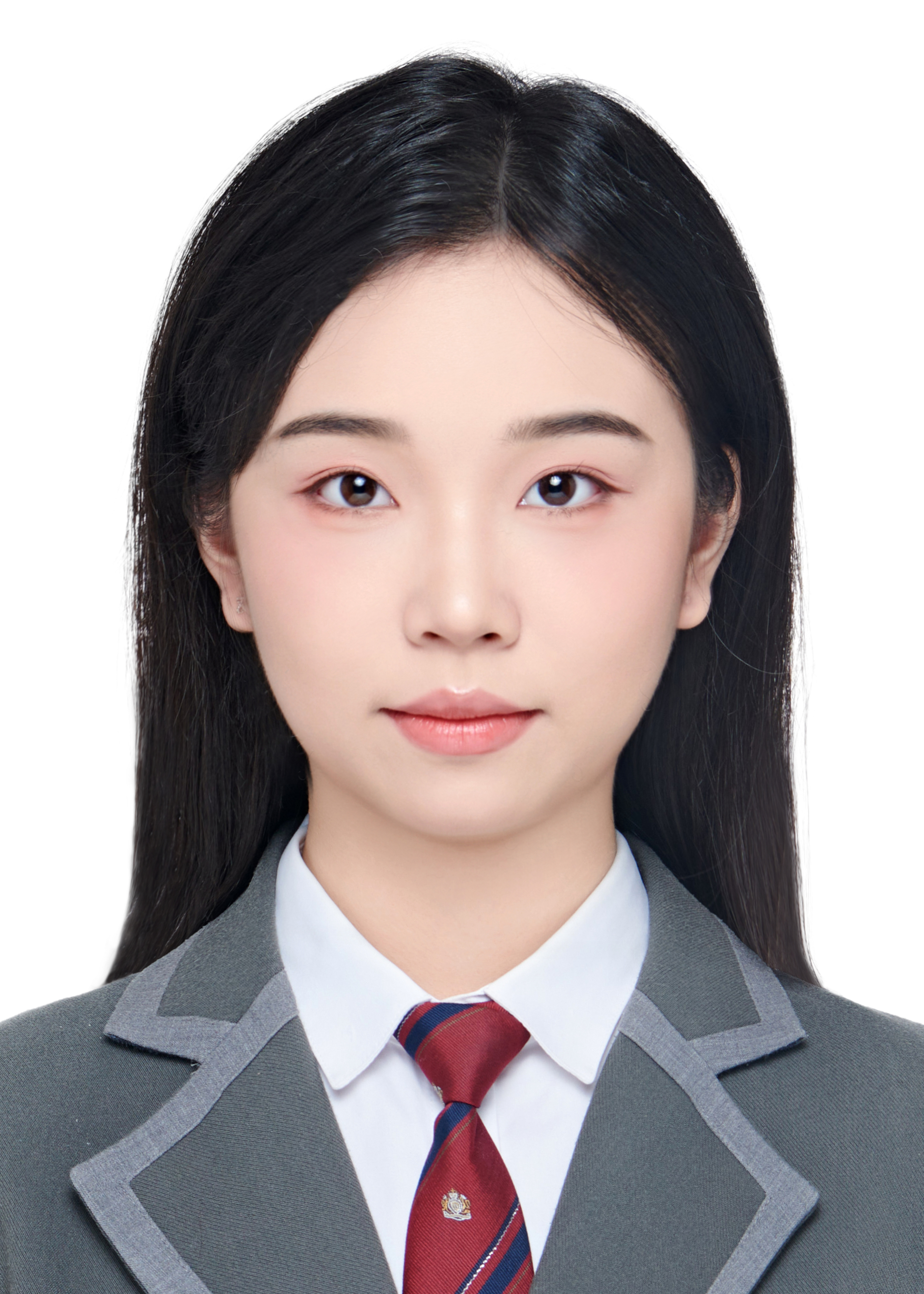}}]{Xinyu Zeng}
 is currently pursuing the M.S. degree in New-Generation Electronic Information Technology with the University of Electronic Science and Technology of China (UESTC), China. Her research interests include trustworthy AI and end-to-end autonomous driving.
\end{IEEEbiography}

\begin{IEEEbiography}[{\includegraphics[width=1in,height=1.25in,clip,keepaspectratio]{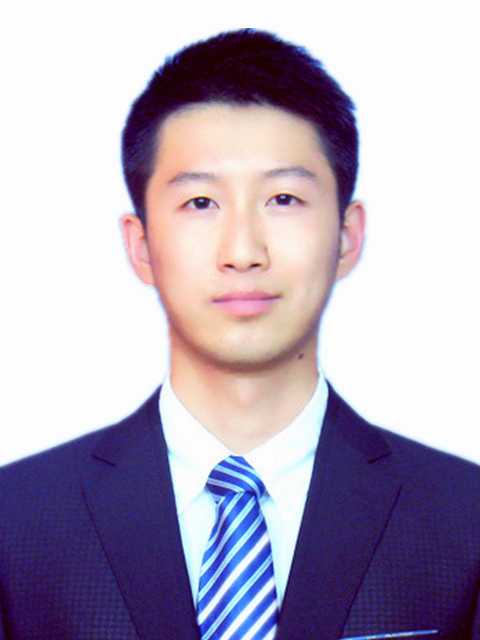}}]{Xiangkun He} (Senior Member, IEEE)
 is currently a UESTC100 Young Professor at the University of Electronic Science and Technology of China. Previously, he was a Research Fellow at Nanyang Technological University, Singapore, and served as a Senior Research Scientist at Huawei Noah's Ark Lab from 2019 to 2021. He earned his Ph.D. in 2019 from the School of Vehicle and Mobility at Tsinghua University. His research interests include reinforcement learning, trustworthy AI, autonomous vehicles, and robotics. He has authored over 60 papers in academic journals and conferences, and has been granted 10 patents. He serves as the Deputy Secretary-General of the Intelligent Driving Professional Committee of the Chinese Association for Artificial Intelligence, as a member of the editorial boards of 6 academic journals, and as a reviewer for over 50 journals and conferences.
\end{IEEEbiography}

\begin{IEEEbiography}[{\includegraphics[width=1in,height=1.25in,clip,keepaspectratio]{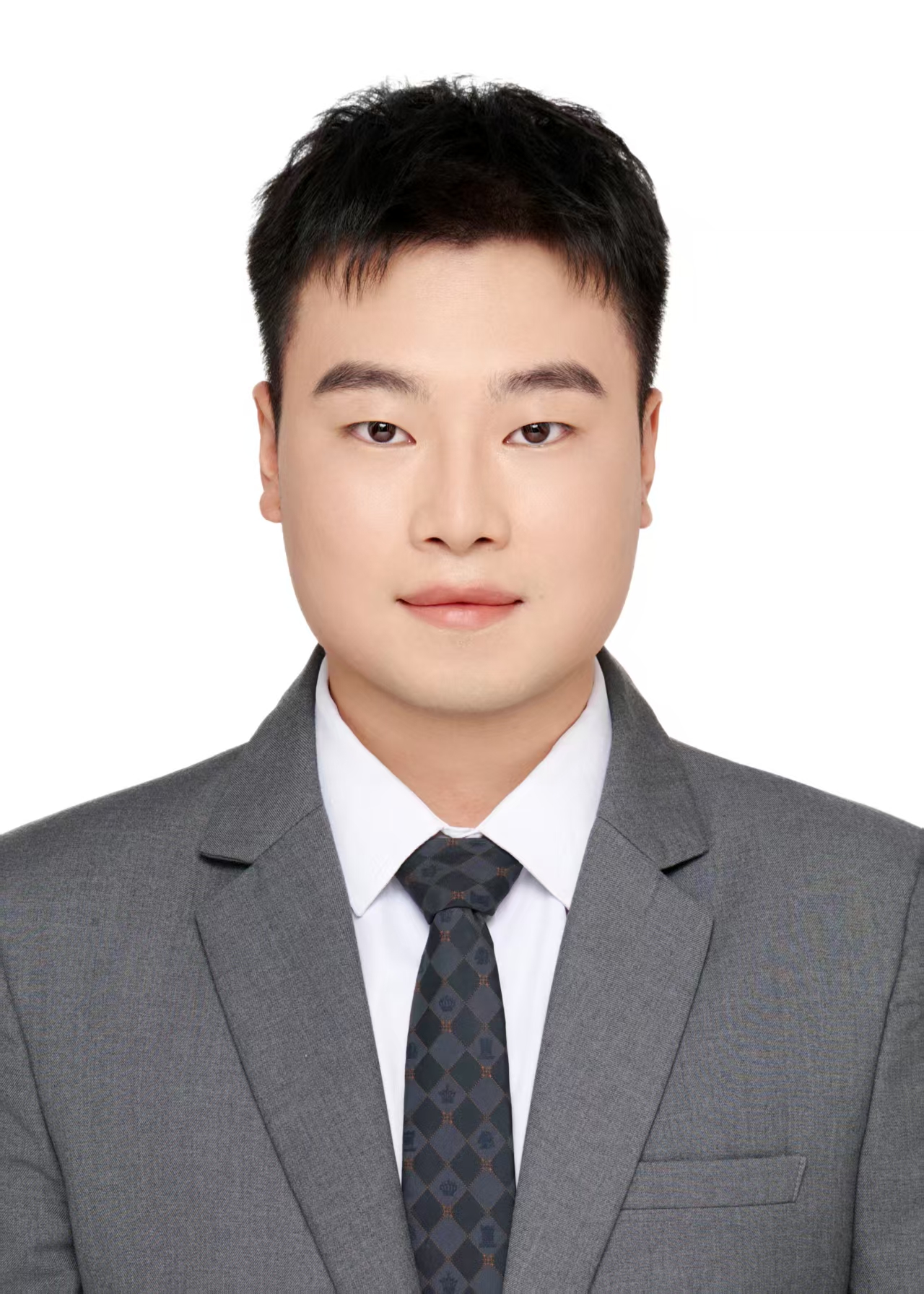}}]{Lei Tao}
 is currently pursuing the M.S. degree in Computer Technology with the University of Electronic Science and Technology of China (UESTC), China. His research focuses on the application of adversarial attack methods in autonomous driving systems.
\end{IEEEbiography}

\begin{IEEEbiography}[{\includegraphics[width=1in,height=1.25in,clip,keepaspectratio]{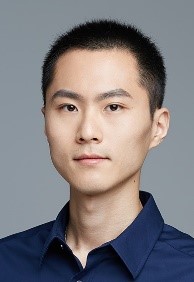}}]{Chen Lv} (Senior Member, IEEE)
	received the PhD
degree from the Department of Automotive Engineering, Tsinghua University, China, in 2016. He
is currently an associate professor with the School
of Mechanical and Aerospace Engineering, Nanyang
Technology University, Singapore. He also holds joint
appointments as the Cluster director of Future Mobility Solutions, ERI@N, Thrust Lead in Smart Mobility
and Delivery, Continental-NTU Corp Lab and Program lead in Next Generation AMR, Schaeffler-NTU
Joint Lab. His research interests include advanced
vehicles and human-machine systems, where he has contributed more than 200
papers and received 12 granted patents in China. Dr. Lv is an associate editor for
IEEE Transactions on Intelligent Transportation Systems, IEEE Transactions
on Vehicular Technology, and IEEE Transactions on Intelligent Vehicles.
\end{IEEEbiography}

\begin{IEEEbiography}[{\includegraphics[width=1in,height=1.25in,clip,keepaspectratio]{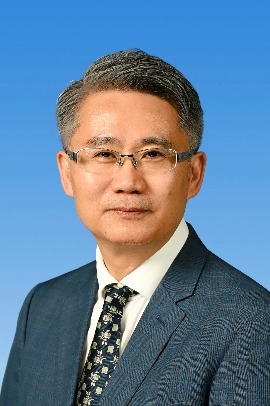}}]{Hong Cheng}  (Senior Member, IEEE) received the Ph.D. degree in pattern recognition and intelligent systems from Xi’an Jiaotong University, Xi’an, China, in 2003. He has been an Associate Professor of Xi’an Jiaotong University since 2005. He was a Post-Doctoral Researcher at the Computer Science School, Carnegie Mellon University, Pittsburgh, PA, USA, from 2006 to 2009. He is currently a Full Professor and the Dean of the School of Mechanical and Electrical Engineering, as well as the Director of the Center for Robotics, at the University of Electronic Science and Technology of China, Chengdu, China. He is also the Founding Director of the Machine Intelligence Institute. His current research interests include computer vision, machine learning, and robotics.
\end{IEEEbiography}

\end{document}